\documentclass[twoside,11pt]{article}

\usepackage{pdflscape}
\usepackage{booktabs}
\usepackage{multirow}
\usepackage{paralist,amsmath, amssymb,bm}
\usepackage{algorithm,algorithmic}
\usepackage{jmlr2e}

\usepackage[colorlinks,
            linkcolor=blue,
            citecolor=blue,
            urlcolor=magenta,
            linktocpage,
            plainpages=false]{hyperref}

\makeatletter
\newcounter{ALC@tempcntr}
\makeatother

\makeatletter
\AtBeginDocument{\Hy@breaklinkstrue}
\makeatother

\newtheorem{thm}{Theorem}

 \newtheorem{ass}{Assumption}

\def \y {\mathbf{y}}

\def \x {\mathbf{x}}
\def \g {\mathbf{g}}

\def \u {\mathbf{u}}
\def \H {\mathcal{H}}
\def \w {\mathbf{w}}
\def \R {\mathbb{R}}

\def \bv {\mathbf{v}}

\def \xb {\bar{\x}}

\def \X {\mathcal{X}}

\newcommand{\tabincell}[2]{\begin{tabular}{@{}#1@{}}#2\end{tabular}}

\DeclareMathOperator*{\DReg}{D-Regret}

\DeclareMathOperator*{\argmin}{argmin}

\def\abovestrut#1{\rule[0in]{0in}{#1}\ignorespaces}
\def\belowstrut#1{\rule[-#1]{0in}{#1}\ignorespaces}

\def\abovespace{\abovestrut{0.20in}}

\def\belowspace{\belowstrut{0.10in}}

\def\abovespaceLarge{\abovestrut{0.30in}}
\def\belowspaceLarge{\belowstrut{0.15in}}

\begin{document}

\title{Revisiting Smoothed Online Learning}

\author{\name Lijun Zhang \email zhanglj@lamda.nju.edu.cn\\
        \name Wei Jiang \email jiangw@lamda.nju.edu.cn\\
        \name Shiyin Lu \email lusy@lamda.nju.edu.cn\\
       \addr National Key Laboratory for Novel Software Technology, Nanjing University, Nanjing 210023, China\\
       \name Tianbao Yang \email tianbao-yang@uiowa.edu \\
 \addr Department of Computer Science,  the University of Iowa, Iowa City, IA 52242, USA
}
\editor{}

\maketitle

\begin{abstract}
In this paper, we revisit the problem of smoothed online learning, in which the online learner suffers both a hitting cost and a switching cost, and target two performance metrics:  competitive ratio and dynamic regret with switching cost. 

To bound the competitive ratio, we assume the hitting cost is known to the learner in each round, and investigate the simple idea of balancing the two costs by an optimization problem.
Surprisingly, we find that minimizing the hitting cost alone is $\max(1, \frac{2}{\alpha})$-competitive for $\alpha$-polyhedral functions and $1 + \frac{4}{\lambda}$-competitive for $\lambda$-quadratic growth functions, both of which improve state-of-the-art results significantly. Moreover, when the hitting cost is both convex and $\lambda$-quadratic growth, we reduce the competitive ratio to $1 + \frac{2}{\sqrt{\lambda}}$  by minimizing the weighted sum of the hitting cost and the switching cost.
 
To bound the dynamic regret with switching cost, we follow the standard setting of online convex optimization, in which the hitting cost is convex but hidden from the learner before making predictions. We modify Ader, an existing algorithm designed for dynamic regret, slightly to take into account the switching cost when measuring the performance. The proposed algorithm, named as Smoothed Ader, attains an optimal $O(\sqrt{T(1+P_T)})$ bound for dynamic regret with switching cost, where $P_T$ is the path-length of the comparator sequence. Furthermore, if the hitting cost is accessible in the beginning of each round, we obtain a similar guarantee without the bounded gradient condition, and establish an $\Omega(\sqrt{T(1+P_T)})$ lower bound to confirm the optimality.
\end{abstract}

\section{Introduction}
Online learning is the process of making a sequence of predictions given knowledge of the answer to previous tasks and possibly additional  information \citep{Online:suvery}. While the traditional online learning aims to make the prediction as accurate as possible, in this paper, we study smoothed online learning (SOL), where the online learner incurs a switching cost for changing its predictions between rounds \citep{NIPS2013_5151}. SOL has received lots of attention recently because in many real-world applications, a change of action usually brings some additional cost. Examples include the dynamic right-sizing for data centers \citep{Dynamic:Data:Center}, geographical load balancing \citep{Geographical:Load:Balance}, real-time electricity pricing \citep{electricity:pricing},  video streaming \citep{video:transport}, spatiotemporal sequence prediction \citep{Spatiotemporal:Prediction}, multi-timescale control \citep{Optimization:timescales}, and thermal management \citep{Management:MPSoCs}.

Specifically, SOL is performed in a sequence of consecutive rounds, where at round $t$ the learner is asked to select a point $\x_t$ from the decision set $\X$, and suffers a hitting cost $f_t(\x_t)$. Depending on the performance metric, the learner \emph{may} be allowed to observe $f_t(\cdot)$ when making decisions, which is different from the traditional online learning in which $f_t(\cdot)$ is revealed to the learner after submitting the decision \citep{bianchi-2006-prediction}. Additionally, the learner also incurs a switching cost $m(\x_t,\x_{t-1})$ for changing decisions between successive rounds. The switching cost $m(\x_t,\x_{t-1})$  could be any distance function, such as the $\ell_2$-norm distance $\|\x_t-\x_{t-1}\|$ and the squared $\ell_2$-norm distance $\|\x_t-\x_{t-1}\|^2/2$ \citep{NIPS2019_8463}. In the literature, there are two performance metrics for SOL: competitive ratio and dynamic regret with switching cost.

Competitive ratio is popular in the community of online algorithms \citep{Online:Competitive:book}. It is defined as the worst-case ratio of the total cost incurred by the online learner and the offline optimal cost:
\begin{equation} \label{eqn:competitive:ratio}
\frac{\sum_{t=1}^T \big( f_t(\x_t) + m(\x_t, \x_{t-1}) \big) }{\min_{\u_0,\u_1,\ldots,\u_T\in \X} \sum_{t=1}^T  \big(f_t (\u_t) + m(\u_t, \u_{t-1})\big)}.
\end{equation}
When focusing on the competitive ratio, the learner can observe $f_t(\cdot)$ before picking $\x_t$. The problem is still nontrivial due to the coupling created by the switching cost. On the other hand, dynamic regret with switching cost is a generalization of dynamic regret---a popular performance metric in the community of online learning \citep{zinkevich-2003-online}. It is defined as the difference between the total cost incurred by the online learner and that of an arbitrary comparator sequence $\u_0,\u_1,\ldots,\u_T \in \X$:
\begin{equation} \label{eqn:dynamic:regret:switching}
\sum_{t=1}^T \big( f_t(\x_t) + m(\x_t, \x_{t-1})\big) - \sum_{t=1}^T  \big(f_t (\u_t) + m(\u_t, \u_{t-1})\big) .
\end{equation}
Different from previous work \citep{SOCO:OBD,NIPS2019_8463}, we did not introduce the minimization operation over $\u_0,\u_1,\ldots,\u_T$ in (\ref{eqn:dynamic:regret:switching}). The reason is that we want to bound (\ref{eqn:dynamic:regret:switching}) by certain regularities of the comparator sequence, such as the path-length
\begin{equation}\label{eqn:path}
P_T(\u_0, \u_1, \ldots, \u_T)=\sum_{t=1}^T \|\u_t - \u_{t-1}\| .
\end{equation}
When focusing on (\ref{eqn:dynamic:regret:switching}), $f_t(\cdot)$ is generally hidden from the learner before submitting $\x_t$. The conditions for bounding the two metrics are very different, so we study competitive ratio and dynamic regret with switching cost separately. To bound the two metrics simultaneously, we refer to \citet{pmlr-v30-Andrew13} and \citet{pmlr-v98-daniely19a}, especially the meta-algorithm in the latter work.

This paper follows the line of research stemmed from online balanced descent (OBD) \citep{SOCO:OBD,OBD:Strongly}. The key idea of OBD is to find an appropriate balance between the hitting cost and the switching cost through iterative projections. It has been shown that OBD and its variants are able to exploit the analytical properties of the hitting cost (e.g., polyhedral, strongly convex) to derive dimension-free competitive ratio. At this point, it would be natural to ask why not use the greedy algorithm, which minimizes the weighted sum of the hitting cost and the switching cost in each round, i.e.,
\begin{equation} \label{eqn:greedy}
\min_{\x\in \X} \quad f_t(\x)+ \gamma m(\x, \x_{t-1})
\end{equation}
to balance the two costs, where $\gamma \geq 0$ is the trade-off parameter. We note that the greedy algorithm is usually treated as the baseline in competitive analysis \citep{Online:Competitive:book}, but its usage for smoothed online learning is quite limited. One result is given by \citet{NIPS2019_8463}, who demonstrate that the greedy algorithm as a special case of Regularized OBD (R-OBD), is optimal for strongly convex functions. Besides, \citet{10.1145/3379484} have analyzed the greedy algorithm with $\gamma=0$, named as the \emph{naive} approach below, for polyhedral functions and quadratic growth functions.

In this paper, we make the following contributions towards understanding the behavior of the greedy algorithm.
\begin{compactitem}
\item For $\alpha$-polyhedral functions, the competitive ratio of the naive approach is $\max(1, \frac{2}{\alpha})$, which is a significant improvement over the $3+\frac{8}{\alpha}$ competitive ratio of OBD \citep{SOCO:OBD} and the $1+\frac{2}{\alpha}$ ratio proved by \citet[Lemma 1]{10.1145/3379484}. When $\alpha>2$, the ratio becomes $1$, indicating that the naive approach is optimal in this scenario.
\item For $\lambda$-quadratic growth functions, the competitive ratio of the naive algorithm is $1 + \frac{4}{\lambda}$, which matches the lower bound of this algorithm \citep[Theorem 5]{NIPS2019_8463}, and is better than the $\max(1+\frac{6}{\lambda},4)$ ratio obtained by \citet[Lemma 1]{10.1145/3379484}.
\item If the hitting cost is both convex and $\lambda$-quadratic growth, the greedy algorithm with $\gamma>0$ attains a $1 + \frac{2}{\sqrt{\lambda}}$ competitive ratio, which demonstrates the advantage of taking the switching cost into considerations. Our $1 + \frac{2}{\sqrt{\lambda}}$ ratio is on the same order as Greedy OBD \citep[Theorem 3]{NIPS2019_8463} but with much smaller constants.
\item  Our analysis of the naive approach and the greedy algorithm is very simple. In contrast,  both OBD and Greedy OBD rely on intricate geometric arguments.
\end{compactitem}

While both OBD  and R-OBD are equipped with sublinear dynamic regret with switching cost, they are unsatisfactory in the following aspects:
\begin{compactitem}
\item The regret  of OBD depends on an upper bound of the path-length instead of the path-length itself \citep[Corollary 11]{SOCO:OBD}, making it nonadaptive.
\item The regret of R-OBD is adaptive but it uses the squared $\ell_2$-norm to measure the switching cost, which is not suitable for general convex functions \citep{NIPS2019_8463}.\footnote{We usually assume the convex function is Lipschitz continuous, and in this case choosing the $\ell_2$-norm distance as the switching cost makes it on the same order as the hitting cost.}
\item Both OBD and R-OBD observe $f_t(\cdot)$ before selecting $\x_t$, which violates the convention of  online learning.
\end{compactitem}

To avoid the above limitations, we demonstrate that a small change of Ader \citep{Adaptive:Dynamic:Regret:NIPS}, which is an existing algorithm designed for dynamic regret, is sufficient to minimize the dynamic regret with switching cost under the setting of online convex optimization \citep{Online:suvery}. Ader runs multiple online gradient descent (OGD)  \citep{zinkevich-2003-online} with different step sizes as expert-algorithms, and uses Hedge \citep{FREUND1997119} as the meta-algorithm to aggregate predictions from experts. The only modification is to incorporate the switching cost into the loss of Hedge. The proposed algorithm, named as Smoothed Ader (SAder), attains the optimal $O(\sqrt{T(1+P_T)})$ dynamic regret, where $P_T$ is the path-length defined in (\ref{eqn:path}). Thus, our regret bound is \emph{adaptive} because it automatically becomes small when the comparators change slowly. Finally, we also investigate the case that the hitting cost is available before predictions, and establish a similar result without the bounded gradient condition. To this end, we design a \emph{lookahead} version of SAder, which chooses the greedy algorithm in (\ref{eqn:greedy}) as the expert and utilizes the cost of the current round in Hedge. To show the optimality of this algorithm, we further establish an $\Omega (\sqrt{T(1+P_T)})$ lower bound under the lookahead setting.
\section{Related Work}
This section reviews related work on smoothed online learning (SOL) and dynamic regret.
\subsection{Smoothed Online Learning}
SOL has been investigated under the setting of multi-armed bandits \citep{SSR:Bandits:Switching,Bandit:Metric:Switching,Dekel:Bandits:Switching,NIPS2017_7000, pmlr-v65-koren17a}, prediction with expert advice \citep{NIPS2013_5151}, and online convex optimization \citep{Dynamic:Data:Center,bansal_et_al:LIPIcs,10.1145/3375788}. In the following, we focus on  smoothed online convex optimization (SOCO).

The early works on SOCO are motivated by real-world applications with switching costs, and focus on designing competitive algorithms in the low-dimensional setting \citep{Dynamic:Data:Center}. In particular, \citet{bansal_et_al:LIPIcs} show that for SOCO on the real line, the competitive ratio can be upper bounded by $2$, which is proved to be optimal \citep{10.1007/978-3-319-89441-6_13}. They also establish a competitive ratio of $3$ under the memoryless setting. In the study of SOCO, it is common to assume that the learner has access to predictions of future hitting costs, and several algorithms \citep{Geographical:Load:Balance,10.1145/2745844.2745854,10.1145/2896377.2901464,Prediction:Switching,NEURIPS2020_a6e4f250} have been developed based on receding horizon control (RHC) \citep{RHC:2005}. In fact, the greedy algorithm in (\ref{eqn:greedy}) can be treated as a variant of RHC. However, previous results for RHC are limited to special problems,  and (\ref{eqn:greedy}) remains under-explored. For example, when the learner can observe the next $W$ hitting costs, \citet{Prediction:Switching} demonstrate that both the competitive ratio and the dynamic regret with switching cost decay exponentially fast with $W$. But their analysis relies on very strong conditions, including strong convexity and smoothness.

One milestone is the online balanced descent (OBD) \citep{SOCO:OBD}, which has dimension-free competitive ratio even when the learner can only observe the hitting cost of the current round. Specifically, OBD iteratively projects the previous point onto a carefully chosen level set of the hitting cost  so as to balance the switching cost and the hitting cost. When the hitting cost is $\alpha$-polyhedral and convex, and the switching cost is the $\ell_2$-norm distance, OBD attains a $3+\frac{8}{\alpha}$ competitive ratio. Furthermore, OBD can also be tuned to control the dynamic regret with switching cost. Let $L$ be an upper bound of the path-length of the comparator sequence, i.e., $P_T(\u_0, \u_1, \ldots, \u_T) \leq L$. \citet[Corollary 11]{SOCO:OBD} have proved that
\begin{equation} \label{eqn:obd:regret}
\sum_{t=1}^T \big( f_t(\x_t) + \|\x_t-\x_{t-1}\|\big) - \min_{P_T(\u_0, \u_1, \ldots, \u_T) \leq L} \sum_{t=1}^T  \big(f_t (\u_t) + \|\u_t- \u_{t-1}\|\big) = O\left(\sqrt{TL}\right)
\end{equation}
leading to sublinear regret when $L=o(T)$. However, there are two limitations: (i) the value of $L$ needs to be given as a parameter; and  (ii) it is nonadaptive because the upper bound depends on $L$ instead of the actual path-length $P_T$.

\begin{table}[t]
\centering\caption{Competitive Ratio of OBD and Related Methods}
\label{tab:results}
\begin{tabular}{c|c|c}
\toprule Algorithm & Assumptions & Competitive Ratio \belowspace\\ \hline
 \multirow{3}{*}{OBD \citep{SOCO:OBD}} & $f_t(\cdot)$ is $\alpha$-polyhedral & \multirow{3}{*}{$3+\frac{8}{\alpha}$} \abovespace  \\
   &  $f_t(\cdot)$ is  convex \\
  & $\X$ is convex  & \belowspace\\ \hline
 The Naive Approach  \abovespace  & \multirow{4}{*}{$f_t(\cdot)$ is $\alpha$-polyhedral} & \multirow{2}{*}{$1+ \frac{2}{\alpha}$}  \\
 \citep{10.1145/3379484}\belowspace & & \\ \cline{1-1} \cline{3-3}
 The Naive Approach  \abovespace  &  &  \multirow{2}{*}{$\max\left(1, \frac{2}{\alpha}\right)$}  \\
 (Theorem~\ref{thm:ratio:poly}) \belowspace & & \\ \hline
 \multirow{3}{*}{Greedy OBD \citep{NIPS2019_8463}} & $f_t(\cdot)$ is $\lambda$-quadratic growth & \multirow{3}{*}{$O\left(\frac{1}{\sqrt{\lambda}}\right)$ as $\lambda \rightarrow 0$}  \abovespace \\
   &  $f_t(\cdot)$ is  quasiconvex \\
  & $\X$ is convex  & \belowspace \\ \hline
  The Naive Approach    \abovespace  & \multirow{4}{*}{$f_t(\cdot)$ is $\lambda$-quadratic growth} & \multirow{2}{*}{$\max\left(1+\frac{6}{\lambda},4\right)$}  \\
 \citep{10.1145/3379484} \belowspace & & \\ \cline{1-1} \cline{3-3}
 The Naive Approach   \abovespace  &  & \multirow{2}{*}{$1 + \frac{4}{\lambda}$}  \\
 (Theorem~\ref{thm:naive:quadratic}) \belowspace & & \\ \hline
 \multirow{3}{*}{\tabincell{c}{The Greedy Algorithm\\  (Theorem~\ref{thm:greedy:convex:quadratic})}} & $f_t(\cdot)$ is $\lambda$-quadratic growth & \multirow{3}{*}{$1+\frac{2}{\sqrt{\lambda}}$ }  \abovespace \\
   &  $f_t(\cdot)$ is  convex \\
 & $\X$ is convex  & \belowspace \\ \hline
 \multirow{2}{*}{OBD \citep{OBD:Strongly}} &  & \multirow{2}{*}{$3+O\left(\frac{1}{\lambda}\right)$} \abovespace  \\
  & $f_t(\cdot)$ is $\lambda$-strongly convex  & \\ \cline{1-1} \cline{3-3}
  The Greedy Algorithm, and & $\X$ is convex & \multirow{2}{*}{$\frac{1}{2}+ \frac{1}{2}\sqrt{1+\frac{4}{\lambda}} $} \abovespace  \\
R-OBD \citep{NIPS2019_8463}  &   & \belowspace\\\hline
 & $f_t(\cdot)$ is $\lambda$-strongly convex \abovespace & \multirow{4}{*}{$4+ 4\sqrt{\frac{2H}{\lambda}}$}   \\
   Constrained OBD  &  $f_t(\cdot)$ is  $H$-smooth \\
   \citep{pmlr-v125-argue20a} & $\min_{x \in \R^d} f_t(\cdot)=0$ & \\
  & $\X$ is convex  &  \belowspace\\
\bottomrule
\end{tabular}
\end{table}

Later, \citet{OBD:Strongly} demonstrate that OBD is $3+O(\frac{1}{\lambda})$-competitive for $\lambda$-strongly convex functions, when the switching cost is set to be the squared $\ell_2$-norm distance. In a subsequent work, \citet[Theorem 4]{NIPS2019_8463} propose Regularized OBD (R-OBD), which improves the competitive ratio to $\frac{1}{2}+ \frac{1}{2}\sqrt{1+\frac{4}{\lambda}}$, matching the lower bound of strongly convex functions exactly  \cite[Theorem 1]{NIPS2019_8463}. R-OBD includes  (\ref{eqn:greedy}) as a special case, which also enjoys the optimal competitive ratio for strongly convex functions. Furthermore, \citet[Theorem 6]{NIPS2019_8463} have analyzed the dynamic regret of R-OBD and the following result can be extracted from that paper
\[
\sum_{t=1}^T \left( f_t(\x_t) + \frac{1}{2}\|\x_t-\x_{t-1}\|^2 \right) -  \sum_{t=1}^T  \left(f_t (\u_t) + \frac{1}{2}\|\u_t- \u_{t-1}\|^2\right) = O\left(\sqrt{T  \sum_{t=1}^T \|\u_t- \u_{t-1}\|^2}\right)
\]
Compared with (\ref{eqn:obd:regret}), this bound is adaptive because the upper bound depends on the switching cost of the comparator sequence. However, it chooses the squared $\ell_2$-norm as the switching cost, which is not suitable for general convex functions, and the proof makes use of the vanishing gradient condition, which may fail because the domain is bounded. When the hitting cost is both quasiconvex and $\lambda$-quadratic growth, \citet[Theorem 3]{NIPS2019_8463} have demonstrated that their Greedy OBD algorithm attains an $O(1/\sqrt{\lambda})$ competitive ratio, as $\lambda \rightarrow  0$.

\citet{10.1145/3379484} have analyzed the naive approach which ignores the switching cost and simply minimizes the hitting cost in each round, i.e., the greedy algorithm with $\gamma=0$. It is a bit surprising that this naive approach is $1+\frac{2}{\alpha}$-competitive for $\alpha$-polyhedral functions and $\max(1+\frac{6}{\lambda},4)$-competitive for $\lambda$-quadratic growth functions, without any convexity assumption \citep[Lemma 1]{10.1145/3379484}. \citet{pmlr-v125-argue20a} have investigated a hybrid setting in which the hitting cost is both $\lambda$-strongly convex and $H$-smooth, but the switching cost is the $\ell_2$-norm distance instead of the squared one. They develop Constrained OBD, and establish a $4+ 4\sqrt{2H/\lambda}$ competitive ratio. However, their analysis relies on a strong condition that the hitting cost is non-negative over the whole space, i.e., $\min_{x \in \R^d} f_t(\cdot)=0$, as opposed to the usual condition $\min_{x \in \X} f_t(\x)=0$.  To facilitate comparisons, we summarize the competitive ratios of OBD, its variants, and the greedy algorithm in Table~\ref{tab:results}.

Finally, we note that SOCO is closely related to convex body chasing (CBC) \citep{Convex:Body,CCBF:2016,NCB:2018,Linear:CNCB,CCCB:Bubeck,Optimal:CNCB}. In this problem, the online learner receives a sequence of convex bodies $\X_1,\ldots,\X_T \subseteq \R^d$ and must select one point from each body, and the goal is to minimize the total movement between consecutive output points. Apparently, we can treat CBC as a special case of SOCO by defining the hitting cost $f_t(\cdot)$ as the indicator function of $\X_t$, which means that the domains of hitting costs are allowed to be different. On the other hand, we can also formulate a $d$-dimensional SOCO problem as a $d+1$-dimensional CBC problem \citep[Proposition 1]{10.1145/3379484}. For the general setting of CBC, the competitive ratio exhibits a polynomial dependence on the dimensionality, and the state-of-the-art result is $O(\min(d, \sqrt{d \log T}))$  \citep{Linear:CCB,Optimal:CCB}, which nearly match the $\Omega(\sqrt{d})$ lower bound \citep{Convex:Body}. Our paper aims to derive dimensionality-independent competitive ratios and sublinear dynamic regret for SOCO, under appropriate conditions.

\subsection{Dynamic Regret}
Recently, dynamic regret has attained considerable interest in the community of online learning \citep{IJCAI:2020:Zhang}. The motivation of dynamic regret is to deal with changing environments, in which the optimal decision may change over time. It is defined as the difference between the cumulative loss of the learner and that of a sequence of comparators $\u_1, \ldots, \u_T \in \X$:
\begin{equation} \label{eqn:dynamic:1}
\DReg(\u_1,\ldots,\u_T) = \sum_{t=1}^T f_t(\x_t)  - \sum_{t=1}^T f_t(\u_t).
\end{equation}
In the general form of dynamic regret,  $\u_1, \ldots, \u_T$ could be an \emph{arbitrary} sequence \citep{zinkevich-2003-online,Dynamic:ICML:13,Adaptive:Dynamic:Regret:NIPS,Problem:Dynamic:Regret,pmlr-v119-cutkosky20a}, and in the restricted form,  they are chosen as the minimizers of online functions, i.e., $\u_t \in \argmin_{\x \in \X} f_t(\x)$ \citep{Dynamic:AISTATS:15,Non-Stationary,Dynamic:2016,Dynamic:Strongly,Dynamic:Regret:Squared,Worst:Dynamic:SSS}. While it is well-known that sublinear dynamic regret is unattainable in the worst case, one can bound the dynamic regret in terms of some regularities of the comparator sequence. An instance is given by \citet{zinkevich-2003-online}, who introduces the notion of path-length defined in (\ref{eqn:path}) to measure the temporal variability of the comparator sequence, and derives an $O(\sqrt{T}(1+P_T))$ bound for the dynamic regret of OGD. Later, \citet{Adaptive:Dynamic:Regret:NIPS} develop adaptive learning for dynamic environment (Ader), which achieves the optimal $O(\sqrt{T(1+P_T)})$ dynamic regret. In this paper, we show that a small change of Ader attains the same bound for dynamic regret with switching cost.  We summarize our  and previous results on dynamic regret  in Table~\ref{tab:results:dynamic:regret}. 

\section{Competitive Ratio}
In this section, we choose competitive ratio as the performance metric. Without loss of generality, we assume the hitting cost is non-negative, since the competitive ratio can only improve if this is not the case.

\subsection{Polyhedral Functions}
We first introduce the definition of polyhedral functions.
\begin{definition} A function $f(\cdot): \X \mapsto \R$ with minimizer $\bv$ is
$\alpha$-polyhedral if
\begin{equation} \label{eqn:polyhedral}
f(\x)- f(\bv) \geq \alpha \|\x-\bv\|,  \  \forall \x  \in \X.
\end{equation}
\end{definition}
We note that polyhedral functions have been used for stochastic network optimization \citep{stochastic:network:opt} and geographical load balancing \citep{Geographical:Load:Balance}.

Following \citet{SOCO:OBD}, we set the switching cost as $m(\x_t,\x_{t-1})=\|\x_t-\x_{t-1}\|$. Intuitively, we may expect that the switching cost should be taken into consideration when making decisions. However, our analysis shows that minimizing the hitting cost alone yields the tightest competitive ratio so far. Specifically, we consider the following naive approach that ignores the switching cost and selects
\begin{equation} \label{eqn:naive}
\x_t = \argmin_{\x\in \X} f_t(\x).
\end{equation}

The theoretical guarantee of (\ref{eqn:naive}) is stated below.
\begin{thm} \label{thm:ratio:poly} Suppose each $f_t(\cdot): \X \mapsto \R$  with minimizer $\bv_t$ is $\alpha$-polyhedral. We have
\[
\sum_{t=1}^T \big( f_t(\x_t) + \|\x_t-\x_{t-1}\| \big)  \leq  \max\left( 1, \frac{2}{\alpha} \right) \sum_{t=1}^T  \big(f_t (\u_t) + \|\u_t - \u_{t-1}\|\big),  \ \forall \u_0, \u_1, \ldots, \u_T \in \X
\]
where we assume $\x_0=\u_0$.
\end{thm}

\begin{landscape}
\begin{table}[h]
\centering\caption{Results on Dynamic Regret with or without Switching Cost. In this table, all the algorithms assume that (i) $f_t(\cdot)$ is  convex and (ii)  $\X$ is bounded, so we omit these two conditions. The condition ``$f_t(\cdot)$ is known/unknown'' is relative to the beginning of the $t$-th round.}
\label{tab:results:dynamic:regret}
\begin{tabular}{@{}c@{\hspace{.5ex}}|@{\hspace{.5ex}}c@{\hspace{.5ex}}|@{\hspace{.5ex}}c@{\hspace{.5ex}}|@{\hspace{.5ex}}c@{}}
\toprule Algorithm & Assumptions & Metric & Upper Bounds \belowspace\\ \hline
OBD  & $f_t(\cdot)$ is known \abovespace  \belowspaceLarge&
  \multirow{2}{*}{$\displaystyle  \sum_{t=1}^T \Big(f_t(\x_t) + \|\x_t-\x_{t-1}\| -   f_t (\u_t) - \|\u_t- \u_{t-1}\| \Big)$} &
  \multirow{2}{*}{$O\left(\sqrt{T L } \right)$}  \\
 \citep{SOCO:OBD} \belowspace & $\sum_{t=1}^T \|\u_t- \u_{t-1}\| \leq L$  &  &\\ \hline
   R-OBD  \abovespaceLarge  & \multirow{2}{*}{$f_t(\cdot)$ is known} &
\multirow{2}{*}{$\displaystyle  \sum_{t=1}^T \Big(f_t(\x_t) + \frac{1}{2}\|\x_t-\x_{t-1}\|^2 -   f_t (\u_t) - \frac{1}{2}\|\u_t- \u_{t-1}\|^2 \Big)$} &
  \multirow{2}{*}{$\displaystyle O\left(\sqrt{T  \sum_{t=1}^T \|\u_t- \u_{t-1}\|^2}\right)$}  \\
\citep{NIPS2019_8463}  \abovespaceLarge  \belowspaceLarge &  &  &
   \\\hline
 OGD \abovespace  \belowspace  &  &  & \multirow{2}{*}{$\displaystyle O\left(\sqrt{T } \sum_{t=1}^T  \|\u_t - \u_{t-1}\| \right)$}  \\
\citep{zinkevich-2003-online}   &   &  \multirow{4}{*}{$\displaystyle  \sum_{t=1}^T \Big(f_t(\x_t) -   f_t(\u_t) \Big)$}  \belowspaceLarge &\\ \cline{1-1}    \cline{4-4}
Ader \abovespace  \belowspace  & $f_t(\cdot)$ is unknown  &    &
  \multirow{6}{*}{$\displaystyle O\left(\sqrt{T \sum_{t=1}^T  \|\u_t - \u_{t-1}\|} \right)$}  \\
\citep{Adaptive:Dynamic:Regret:NIPS}  \belowspace & Gradients are bounded  &  &\\ \cline{1-1} \cline{3-3}
SAder \abovespace  \belowspace  &   &
  \multirow{4}{*}{$\displaystyle \sum_{t=1}^T \Big( f_t(\x_t) + \|\x_t -\x_{t-1}\|  -  f_t(\u_t)\Big)$} &  \\
 (Theorem \ref{thm:ader}) \belowspace&   &  &\\ \cline{1-2}
  Lookahead SAder   \abovespace  \belowspace & \multirow{2}{*}{$f_t(\cdot)$ is known}  &
   &  \\
  (Theorem \ref{thm:lookahead:ader}) \belowspace &    & &\\
\bottomrule
\end{tabular}
\end{table}
\end{landscape}

\textbf{Remark:} Our $\max(1, \frac{2}{\alpha})$ competitive ratio is much better than the $3+\frac{8}{\alpha}$ ratio of OBD \citep{SOCO:OBD}, and also better than the  $1+\frac{2}{\alpha}$ ratio established by \citet{10.1145/3379484} for (\ref{eqn:naive}).  When $\alpha>2$, the ratio becomes $1$, indicating that the naive approach is optimal in this scenario.  Furthermore, the proof of Theorem~\ref{thm:ratio:poly} is much simpler than that of OBD, and refines that of \citet[Lemma 1]{10.1145/3379484}.

To examine the role of the switching cost, we have tried to analyze the greedy algorithm (\ref{eqn:greedy}) with $\gamma >0$, but failed to obtain better ratios.\footnote{In the early version of this manuscript, we have proved a $1+ \frac{2}{\alpha}$ competitive ratio for the greedy algorithm (\ref{eqn:greedy}) with $\gamma >0$ when dealing with $\alpha$-polyhedral functions.} So, the current knowledge suggests that there is no need to consider the switching cost when facing polyhedral functions.  It is unclear whether this is an artifact of our analysis or an inherent property, and will be investigated in the future.

\subsection{Quadratic Growth Functions}
In this section, we consider the quadratic growth condition.
\begin{definition} A function $f(\cdot): \X \mapsto \R$ with minimizer $\bv$ is
$\lambda$-quadratic growth if
\begin{equation} \label{eqn:quadratic:growth}
f(\x)- f(\bv) \geq \frac{\lambda}{2} \|\x-\bv\|^2,  \  \forall \x  \in \X.
\end{equation}
\end{definition}
The quadratic growth condition has been exploited by the optimization community \citep{moor.2017.0889,Necoara2019} to establish linear convergence, and this condition is weaker than  strong convexity \citep{COLT:Hazan:2011}.

Following \citet{NIPS2019_8463}, we set the switching cost as $m(\x_t,\x_{t-1})=\|\x_t-\x_{t-1}\|^2/2$. We also consider the naive approach in (\ref{eqn:naive}) and have the following theoretical guarantee.
\begin{thm} \label{thm:naive:quadratic} Suppose each $f_t(\cdot): \X \mapsto \R$  with minimizer $\bv_t$ is $\lambda$-quadratic growth. We have
\[
\begin{split}\sum_{t=1}^T \left( f_t(\x_t) + \frac{1}{2}\|\x_t-\x_{t-1}\|^2 \right)  \leq \left( 1 + \frac{4}{\lambda}\right) \sum_{t=1}^T  \left(f_t (\u_t) + \frac{1}{2} \|\u_t - \u_{t-1}\|^2\right),
 \end{split}
\]
for all $\u_0, \u_1, \ldots, \u_T \in \X$, where we assume $\x_0=\u_0$.
\end{thm}

\textbf{Remark:} The above theorem implies that the naive approach achieves a competitive ratio of $1 + \frac{4}{\lambda}$, which matches the lower bound of this algorithm \citep[Theorem 5]{NIPS2019_8463}. Furthermore, it is also much better than the $\max(1+\frac{6}{\lambda},4 )$ ratio established by \citet{10.1145/3379484} for (\ref{eqn:naive}). Similar to the case of polyhedral functions, it seems safe to ignore the  switching cost here.\footnote{In the early version of this manuscript, we have proved a $1 + \frac{2(\sqrt{1+\lambda}+1)}{\lambda} $ competitive ratio for the greedy algorithm (\ref{eqn:greedy}) with $\gamma >0$ when dealing with $\lambda$-quadratic growth functions.}

\subsection{Convex and Quadratic Growth Functions}
When $f_t(\cdot)$ is both quasiconvex and $\lambda$-quadratic growth, \citet{NIPS2019_8463} have established an $O(1/\sqrt{\lambda})$ competitive ratio for Greedy OBD. Inspired by this result, we introduce convexity to further improve the competitive ratio. In this case, the switching cost plays a role in deriving tighter competitive ratios.  Specifically, we choose the greedy algorithm with $\gamma >0$ to select $\x_t$, i.e.,
\begin{equation} \label{eqn:2}
\x_t = \argmin_{\x\in \X}\left(  f_t(\x)+ \frac{\gamma}{2} \|\x-\x_{t-1}\|^2 \right).
\end{equation}
The theoretical guarantee of (\ref{eqn:2}) is stated below.

\begin{thm} \label{thm:greedy:convex:quadratic} Suppose the domain $\X$ is convex, and each $f_t(\cdot): \X \mapsto \R$  with minimizer $\bv_t$ is $\lambda$-quadratic growth and convex. By setting
\[
\gamma=\frac{\lambda}{\lambda + \sqrt{\lambda}},
\]
 we have
\[
\begin{split}
\sum_{t=1}^T \left( f_t(\x_t) + \frac{1}{2}\|\x_t-\x_{t-1}\|^2 \right) \leq  \left( 1 + \frac{2}{\sqrt{\lambda}}\right) \sum_{t=1}^T  \left(f_t (\u_t) + \frac{1}{2} \|\u_t - \u_{t-1}\|^2\right),
 \end{split}
\]
for all $\u_0, \u_1, \ldots, \u_T \in \X$, where we assume $\x_0=\u_0$.
\end{thm}
\textbf{Remark:} The above theorem shows that the competitive ratio is improved to $1 + \frac{2}{\sqrt{\lambda}}$ under the additional convexity condition. According to the $\frac{1}{2}+ \frac{1}{2}\sqrt{1+\frac{4}{\lambda}}$ lower bound of the competitive ratio of strongly convex functions \citep[Theorem 1]{NIPS2019_8463}, the ratio in Theorem~\ref{thm:greedy:convex:quadratic} is optimal up to constant factors. Compared with Greedy OBD \citep{NIPS2019_8463}, our assumption is slightly stronger, since we require convexity instead of quasiconvexity. However, our algorithm and analysis are much simpler, and the constants in our bound are much smaller.

\section{Dynamic Regret with Switching Cost}
When considering dynamic regret with switching cost,  we adopt the common assumptions of online convex optimization (OCO) \citep{Online:suvery}.
\begin{ass}\label{ass:0} All the functions $f_t$'s are convex over their domain $\X$.
\end{ass}
\begin{ass}\label{ass:1} The gradients of all functions are bounded by $G$, i.e.,
\begin{equation}\label{eqn:grad}
\max_{\x \in \X}\|\nabla f_t(\x)\| \leq G, \ \forall t \in [T].
\end{equation}
\end{ass}
\begin{ass}\label{ass:2} The diameter of the domain $\X$ is bounded by $D$, i.e.,
\begin{equation}\label{eqn:domain}
\max_{\x, \x' \in \X} \|\x -\x'\| \leq D.
\end{equation}
\end{ass}
Assumption~\ref{ass:1} implies that the hitting cost is Lipschitz continuous, so it is natural to set the switching cost as $m(\x_t,\x_{t-1})=\|\x_t-\x_{t-1}\|$.

\subsection{The Standard Setting}
We first follow the standard setting of OCO in which the learner can not observe the hitting cost when making predictions, and develop an algorithm based on Ader \citep{Adaptive:Dynamic:Regret:NIPS}. 
Specifically, we demonstrate that a small change of Ader, which modifies the loss of the meta-algorithm to take into account the switching cost of experts, is sufficient to minimize the dynamic regret with switching cost. Our proposed method is named as Smoothed Ader (SAder), and stated below.

\begin{algorithm}[t]
\caption{SAder: Meta-algorithm}
\begin{algorithmic}[1]
\REQUIRE A step size $\beta$, and a set $\H$ containing  step sizes for experts
\STATE Activate a set of experts $\{E^\eta| \eta \in \H \}$ by invoking Algorithm~\ref{alg:2} for each step size $\eta \in \H$
\STATE Sort step sizes in ascending order $\eta_1 \leq \eta_2 \leq \cdots \leq \eta_N$, and set $w_1^{\eta_i}=\frac{C}{i(i+1)}$
\FOR{$t=1,\ldots,T$}
\STATE Receive $\x_t^\eta$ from each expert $E^\eta$
\STATE Output the weighted average $\x_t = \sum_{\eta \in \H} w_t^\eta \x_t^\eta$
\STATE Observe the loss function $f_t(\cdot)$
\STATE Update the weight of each expert by (\ref{eqn:meta:update})
\STATE Send gradient $\nabla f_t(\x_t)$ to each expert $E^\eta$
\ENDFOR
\end{algorithmic}
\label{alg:1}
\end{algorithm}

\paragraph{Meta-algorithm} The meta-algorithm is similar to that of Ader \citep[Algorithm 3]{Adaptive:Dynamic:Regret:NIPS}, and summarized in  Algorithm~\ref{alg:1}. The inputs of the meta-algorithm are its own step size $\beta$, and a set $\H$ of step sizes for experts. In Step 1, we active a set of experts $\{E^\eta |\eta \in \H\}$ by invoking the expert-algorithm for each $\eta \in \H$. In Step 2, we set the initial weight of each expert. Let $\eta_i$ be the $i$-th smallest step size in $\H$. The weight of $E^{\eta_i}$ is chosen as
\begin{equation}\label{eqn:initial:weight}
w_1^{\eta_i}=\frac{C}{i(i+1)}, \textrm{ and } C=1+\frac{1}{|\H|}.
\end{equation}
In each round, the meta-algorithm receives a set of predictions $\{\x_t^\eta|\eta \in \H\}$ from all experts (Step 4), and outputs the weighted average (Step 5):
\[
\x_t = \sum_{\eta \in \H} w_t^\eta \x_t^\eta
\]
where $w_t^\eta$ is the weight assigned to expert $E^\eta$. After observing the loss function, the weights of experts are updated according to the  exponential weighting scheme (Step 7) \citep{bianchi-2006-prediction}:
\begin{equation} \label{eqn:meta:update}
w_{t+1}^\eta = \frac{w_{t}^\eta e^{-\beta \ell_t(\x_t^\eta)} }{\sum_{\eta \in \H} w_{t}^\eta e^{-\beta \ell_t(\x_t^\eta)}}
\end{equation}
where
\begin{equation} \label{eqn:meta:loss:function}
\ell_t(\x_t^\eta) = \langle \nabla f_t(\x_t), \x_t^\eta -\x_t \rangle + \|\x_t^\eta - \x_{t-1}^\eta\|.
\end{equation}
When $t=1$, we set $\x_0^\eta=0$, for all $\eta \in \H$. As can be seen from (\ref{eqn:meta:loss:function}), we incorporate the switching cost $\|\x_t^\eta - \x_{t-1}^\eta\|$ of expert $E^\eta$ to measure its performance. This is the \emph{only} modification made to Ader. In the last step, we send the gradient $\nabla f_t(\x_t)$ to each expert $E^\eta$ so that they can update their own predictions.

\begin{algorithm}[t]
\caption{SAder: Expert-algorithm}
\begin{algorithmic}[1]
\REQUIRE The step size $\eta$
\STATE Let $\x_1^\eta$ be any point in $\X$
\FOR{$t=1,\ldots,T$}
\STATE Submit $\x_t^\eta$ to the meta-algorithm
\STATE Receive gradient $\nabla f_t(\x_t)$ from the meta-algorithm
\STATE
\[
\x_{t+1}^\eta= \Pi_{\X}\big[\x_t^\eta - \eta \nabla f_t(\x_t)\big]
\]
\ENDFOR
\end{algorithmic}
\label{alg:2}
\end{algorithm}
\paragraph{Expert-algorithm} The expert-algorithm is the same as that of Ader \citep[Algorithm 4]{Adaptive:Dynamic:Regret:NIPS}, which is OGD over the linearized loss or the surrogate loss
\begin{equation} \label{eqn:surrogate}
s_t(\x)=\langle \nabla f_t(\x_t), \x -\x_t \rangle .
\end{equation}
For the sake of completeness, we present its procedure in Algorithm~\ref{alg:2}. The input of the expert is its step size $\eta$. In Step 3 of Algorithm~\ref{alg:2}, each expert submits its prediction $\x_t^\eta$ to the meta-algorithm, and receives the gradient $\nabla f_t(\x_t)$ in Step 4. Then, in Step 5, it performs gradient descent
\[
\x_{t+1}^\eta= \Pi_{\X}\big[\x_t^\eta - \eta \nabla f_t(\x_t)\big]
\]
to get the prediction for the next round. Here, $\Pi_{\X}[\cdot]$ denotes the projection onto the nearest point in $\X$.

We have the following theoretical guarantee.
\begin{thm} \label{thm:ader} Set
\begin{equation} \label{eqn:def:H}
\H=\left\{ \left. \eta_i=2^{i-1} \sqrt{\frac{D^2}{T (G^2+2G)}} \right  | i=1,\ldots, N\right\}
\end{equation}
where
\[
N= \left\lceil \frac{1}{2} \log_2 (1+2T) \right\rceil +1, \textrm{ and } \beta= \frac{2}{(2G+1) D} \sqrt{\frac{2 }{5 T}}
\]
in Algorithm~\ref{alg:1}. Under Assumptions~\ref{ass:0}, \ref{ass:1} and \ref{ass:2}, for \emph{any} comparator sequence $\u_0, \u_1,\ldots,\u_T \in \X$, SAder satisfies
\begin{align}
& \sum_{t=1}^T \Big( f_t(\x_t) + \|\x_t -\x_{t-1}\|\Big)-  \sum_{t=1}^T  f_t(\u_t) \label{eqn:sader:metric} \\
\leq & \frac{3}{2}\sqrt{T (G^2+2G)\left(D^2+ 2 D  \sum_{t=1}^T  \|\u_t - \u_{t-1}\| \right)}+ (2G+1) D\sqrt{\frac{5  T }{8}} \left[1+ 2\ln (k+1)\right] \nonumber \\
=&O\left(\sqrt{T(1+P_T)} + \sqrt{T}(1+\log \log P_T)\right)=O\left(\sqrt{T(1+P_T)} \right) \nonumber
\end{align}
where we define $\x_0=0$, and
\begin{equation} \label{eqn:def:k}
k = \left\lfloor \frac{1}{2} \log_2 \left(1+ \frac{2P_T}{D} \right) \right \rfloor+1.
\end{equation}
\end{thm}
\textbf{Remark:}  Theorem~\ref{thm:ader} shows that SAder attains an $O(\sqrt{T(1+P_T)})$ bound for dynamic regret with switching cost, which is on the same order as that of Ader for dynamic regret. 
From the $\Omega(\sqrt{T(1+P_T)})$ lower bound of dynamic regret \citep[Theorem 2]{Adaptive:Dynamic:Regret:NIPS}, we know that our upper bound is optimal up to constant factors. Compared with the regret bound of OBD in (\ref{eqn:obd:regret}) \citep{SOCO:OBD}, the advantage of SAder is that its regret depends on the path-length $P_T$ directly, and thus  becomes tighter when focusing on comparator sequences with smaller path-lengths. Finally, note that in (\ref{eqn:sader:metric}), we did not minus the switching cost of the comparator sequence, i.e., $P_T$, that is because it is always smaller than $\sqrt{D T(1+P_T)}$ and does not affect the order.

\begin{algorithm}[t]
\caption{Lookahead SAder: Meta-algorithm}
\begin{algorithmic}[1]
\REQUIRE A step size $\beta$, and a set $\H$ containing step sizes for experts
\STATE Activate a set of experts $\{E^\eta| \eta \in \H \}$ by invoking Algorithm~\ref{alg:4} for each step size $\eta \in \H$
\STATE Sort step sizes in ascending order $\eta_1 \leq \eta_2 \leq \cdots \leq \eta_N$, and set $w_0^{\eta_i}=\frac{C}{i(i+1)}$
\FOR{$t=1,\ldots,T$}
\STATE Observe the loss function $f_t(\cdot)$ and send it to each expert $E^\eta$
\STATE Receive $\x_t^\eta$ from each expert $E^\eta$
\STATE Update the weight of each expert by (\ref{eqn:lookahead:hedge:pro})
\STATE Output the weighted average $\x_t = \sum_{\eta \in \H} w_t^\eta \x_t^\eta$
\ENDFOR
\end{algorithmic}
\label{alg:3}
\end{algorithm}

\subsection{The Lookahead Setting}
It is interesting to investigate whether we can do better if the hitting cost is available before predictions.  In this case, we propose a \emph{lookahead} version of SAder, and demonstrate that the regret bound remains on the same order, but Assumption~\ref{ass:1} can be dropped. That is, the gradient of the function could be unbounded, and thus the function could also be unbounded.

\paragraph{Meta-algorithm} We design a lookahead version of Hedge, and summarize it in Algorithm~\ref{alg:3}. Compared with Algorithm~\ref{alg:1}, we make the following modifications.
\begin{compactitem}
\item In the $t$-th round, the meta-algorithm first sends $f_t(\cdot)$ to all experts so that they can also benefit from the prior knowledge of $f_t(\cdot)$ (Step 4).
\item After receiving the prediction from experts (Step 5), the meta-algorithm makes use of $f_t(\cdot)$ to determine the weights of experts (Step 6):
\begin{equation} \label{eqn:lookahead:hedge:pro}
w_{t}^\eta = \frac{w_{t-1}^\eta e^{-\beta \ell_t(\x_{t}^\eta)} }{\sum_{\eta \in \H} w_{t-1}^\eta e^{-\beta \ell_t(\x_t^\eta)}}
\end{equation}
where $\ell_t(\x_t^\eta)$ is defined in (\ref{eqn:meta:loss:function}).
\end{compactitem}

\paragraph{Expert-algorithm} To exploit the hitting cost of the current round, we choose an instance of the greedy algorithm in (\ref{eqn:greedy}) as the expert-algorithm, and summarize it in Algorithm~\ref{alg:4}. The input of the expert is its step size $\eta$. After receiving $f_t(\cdot)$ (Step 2), the expert solves the following optimization problem to obtain $\x_t^\eta$ (Step 3):
\begin{equation} \label{eqn:lookhead:expert}
\min_{\x\in \X} \quad f_t(\x)+ \frac{1}{2 \eta} \|\x -\x_{t-1}^\eta\|^2 .
\end{equation}

\begin{algorithm}[t]
\caption{Lookahead SAder: Expert-algorithm}
\begin{algorithmic}[1]
\REQUIRE The step size $\eta$
\FOR{$t=1,\ldots,T$}
\STATE Receive the loss $f_t(\cdot)$ from the meta-algorithm
\STATE Solve the optimization problem in (\ref{eqn:lookhead:expert}) to obtain $\x_t^\eta$
\STATE Submit $\x_t^\eta$ to the meta-algorithm
\ENDFOR
\end{algorithmic}
\label{alg:4}
\end{algorithm}
We have the following theoretical guarantee of the lookahead SAder.
\begin{thm} \label{thm:lookahead:ader} Set
\begin{equation} \label{eqn:def:H:New}
\H=\left\{ \left. \eta_i=2^{i-1} \sqrt{\frac{D^2}{T}} \right  | i=1,\ldots, N\right\}
\end{equation}
where
\[
N= \left\lceil \frac{1}{2} \log_2 (1+2T) \right\rceil +1, \textrm{ and } \beta=  \frac{1}{D}  \sqrt{\frac{2}{T}}
\]
in Algorithm~\ref{alg:3}. Under Assumptions~\ref{ass:0} and \ref{ass:2}, for \emph{any} comparator sequence $\u_0, \u_1,\ldots,\u_T \in \X$, the lookahead SAder satisfies
\[
\begin{split}
& \sum_{t=1}^T \Big( f_t(\x_t) + \|\x_t -\x_{t-1}\|\Big)-  \sum_{t=1}^T  f_t(\u_t)\\
\leq & \frac{3}{2}\sqrt{T (D^2+ 2 D \sum_{t=1}^T  \|\u_t - \u_{t-1}\|)}+ D \sqrt{ \frac{T}{2}}  \left[1+ 2\ln (k+1)\right]\\
=&O\left(\sqrt{T(1+P_T)} + \sqrt{T}(1+\log \log P_T)\right)=O\left(\sqrt{T(1+P_T)} \right)
\end{split}
\]
where  $\x_0=0$, and $k$ is defined in (\ref{eqn:def:k}).
\end{thm}
\textbf{Remark:}  Similar to SAder, the lookahead SAder also achieves an $O(\sqrt{T(1+P_T)})$ bound for dynamic regret with switching cost. In the lookahead setting, we do not need Assumption~\ref{ass:1} any more, and the constants in Theorem~\ref{thm:lookahead:ader} are independent from $G$.

To show the optimality of Theorem~\ref{thm:lookahead:ader}, we provide the lower bound of dynamic regret with switching cost under the lookahead setting.
\begin{thm} \label{thm:lower} For any online algorithm with lookahead ability and any $\tau \in [0, TD]$, there exists a sequence of functions $f_1,\ldots,f_T$ and a sequence of comparators $\u_1,\ldots,\u_T$ satisfying Assumptions~\ref{ass:0} and \ref{ass:2} such that (i) the path-length of  $\u_1,\ldots,\u_T$ is at most $\tau$ and (ii) the dynamic regret with switching cost w.r.t.~$\u_1,\ldots,\u_T$ is at least $\Omega(\sqrt{T (D^2 + D \tau)})$.
\end{thm}
\textbf{Remark:}  The above theorem indicates an $\Omega(\sqrt{T(1+P_T)})$ lower bound, which implies that the lookahead SAder is optimal up to constant factors. Thus, even in the lookahead setting, it is impossible to improve the $O(\sqrt{T(1+P_T)})$ upper bound.

\section{Analysis}
In this section, we present the analysis of main theorems, and defer the proofs of supporting lemmas to appendices.

\subsection{Proof of Theorem~\ref{thm:ratio:poly}}
Recall that $\x_t$ is the minimizer of $f_t(\cdot)$, which is $\alpha$-polyhedral. When $t \geq 2$, we have
\[
\begin{split}
 & f_t(\x_t) + \|\x_t - \x_{t-1}\|  \\
\leq & f_t(\x_t) + \|\x_t - \u_{t}\|  + \|\u_{t} -\u_{t-1}\|  + \|\u_{t-1} - \x_{t-1}\|\\
\overset{(\ref{eqn:polyhedral})}{\leq} &   f_t(\x_t) + \frac{1}{\alpha} \big( f_t(\u_{t}) - f_t(\x_t) \big)  + \frac{1}{\alpha} \big( f_{t-1}(\u_{t-1}) - f_{t-1}(\x_{t-1}) \big)  + \|\u_{t} -\u_{t-1}\|.
\end{split}
\]
For $t=1$, we have
\[
\begin{split}
 & f_t(\x_1) + \|\x_1 - \x_{0}\|  \\
\leq & f_1(\x_1) + \|\x_1 - \u_{1}\|  + \|\u_{1} -\u_{0}\|  + \|\u_{0} - \x_{0}\|\\
=& f_1(\x_1) + \|\x_t - \u_{1}\|  + \|\u_{1} -\u_{0}\| \\
\overset{(\ref{eqn:polyhedral})}{\leq} &   f_1(\x_1) + \frac{1}{\alpha} \big( f_1(\u_{1}) - f_1(\x_1) \big)  + \|\u_{1} -\u_{0}\|.
\end{split}
\]

Summing over all the iterations, we have
\begin{equation} \label{eqn:naive:sum}
\begin{split}
 & \sum_{t=1}^T  \big( f_t(\x_t) + \|\x_t - \x_{t-1}\| \big)  \\
\leq & \sum_{t=1}^T f_t(\x_t) + \frac{1}{\alpha} \sum_{t=1}^T  \big( f_t(\u_{t}) - f_t(\x_t) \big)  + \frac{1}{\alpha} \sum_{t=2}^T  \big( f_{t-1}(\u_{t-1}) - f_{t-1}(\x_{t-1}) \big)  + \sum_{t=1}^T  \|\u_{t} -\u_{t-1}\| \\
\leq & \sum_{t=1}^T f_t(\x_t) + \frac{2}{\alpha} \sum_{t=1}^T  \big( f_t(\u_{t}) - f_t(\x_t) \big)  + \sum_{t=1}^T  \|\u_{t} -\u_{t-1}\| \\
= & \frac{2}{\alpha}\sum_{t=1}^T  f_t(\u_{t}) + \sum_{t=1}^T  \|\u_{t} -\u_{t-1}\|  +  \sum_{t=1}^T  \left( 1- \frac{2}{\alpha}\right) f_t(\x_t).
\end{split}
\end{equation}
where the second inequality follows from the fact that $f_T(\x_T) \leq f_T(\u_T)$.

Thus, if $\alpha \geq 2$, we have
\begin{equation} \label{eqn:naive:c1}
\begin{split}
 & \sum_{t=1}^T  \big( f_t(\x_t) + \|\x_t - \x_{t-1}\| \big)  \\
\overset{(\ref{eqn:naive}), (\ref{eqn:naive:sum})}{\leq} & \frac{2}{\alpha} \sum_{t=1}^T  f_t(\u_{t}) + \sum_{t=1}^T  \|\u_{t} -\u_{t-1}\|  +  \sum_{t=1}^T  \left( 1- \frac{2}{\alpha}\right) f_t(\u_t) \\
\leq &  \sum_{t=1}^T  \big( f_t(\u_{t}) + \|\u_{t} -\u_{t-1}\|  \big)  \\
\end{split}
\end{equation}
which implies the naive algorithm is $1$-competitive. Otherwise, we have
\begin{equation} \label{eqn:naive:c2}
\begin{split}
 & \sum_{t=1}^T  \big( f_t(\x_t) + \|\x_t - \x_{t-1}\| \big)  \\
\overset{(\ref{eqn:naive:sum})}{\leq}  & \frac{2}{\alpha} \sum_{t=1}^T f_t(\u_{t}) + \sum_{t=1}^T  \|\u_{t} -\u_{t-1}\|  \leq \frac{2}{\alpha} \sum_{t=1}^T  \big( f_t(\u_{t}) + \|\u_{t} -\u_{t-1}\| \big).
\end{split}
\end{equation}
We complete the proof by combining (\ref{eqn:naive:c1}) and (\ref{eqn:naive:c2}).

\subsection{Proof of Theorem~\ref{thm:naive:quadratic}}
We will make use of the following basic inequality of squared $\ell_2$-norm \citep[Lemma 12]{NIPS2019_8463}.
\begin{equation}\label{eqn:inequality:squared}
\|\x+\y\|^2 \leq (1+\rho) \|\x\|^2 + \left( 1+ \frac{1}{\rho}\right) \|\y\|^2, \ \forall \rho > 0.
\end{equation}

When $t \geq 2$, we have
\[
\begin{split}
 & f_t(\x_t) + \frac{1}{2}\|\x_t - \x_{t-1}\|^2  \\
\overset{(\ref{eqn:inequality:squared})}{\leq} & f_t(\x_t) +  \frac{1+\rho}{2} \|\u_{t} -\u_{t-1}\|^2 + \frac{1}{2}\left( 1+ \frac{1}{\rho}\right) \|\x_t - \x_{t-1} -\u_{t} +\u_{t-1}\|^2 \\
\overset{(\ref{eqn:inequality:squared})}{\leq} & f_t(\x_t) + \frac{1+\rho}{2}  \|\u_{t} -\u_{t-1}\|^2 + \left( 1+ \frac{1}{\rho}\right)  \big( \|\u_{t} -\x_t \|^2 + \|\u_{t-1} -\x_{t-1}\|^2 \big) \\
\overset{(\ref{eqn:quadratic:growth})}{\leq} & f_t(\x_t) + \frac{1+\rho}{2}  \|\u_{t} -\u_{t-1}\|^2 + \frac{2}{\lambda} \left( 1+ \frac{1}{\rho}\right)  \big( f_t(\u_{t})- f_t(\x_t)  + f_{t-1}(\u_{t-1})-f_{t-1}(\x_{t-1}) \big) .
\end{split}
\]
For $t=1$, we have
\[
 f_t(\x_1) + \frac{1}{2} \|\x_1 - \x_{0}\|^2  \overset{(\ref{eqn:inequality:squared}),(\ref{eqn:quadratic:growth}) }{\leq}    f_1(\x_1) +\frac{1+\rho}{2}  \|\u_{1} -\u_{0}\|^2 + \frac{2}{\lambda} \left( 1+ \frac{1}{\rho}\right)  \big( f_1(\u_{1})- f_1(\x_1)   \big).
\]

Summing over all the iterations, we have
\begin{equation} \label{eqn:naive:sum:quadratic}
\begin{split}
 & \sum_{t=1}^T  \left( f_t(\x_t) + \frac{1}{2} \|\x_t - \x_{t-1}\|^2 \right)  \\
\leq & \sum_{t=1}^T f_t(\x_t)+\frac{1+\rho}{2}   \sum_{t=1}^T  \|\u_{t} -\u_{t-1}\|^2  + \frac{2}{\lambda} \left( 1+ \frac{1}{\rho}\right)  \sum_{t=1}^T  \big( f_t(\u_{t}) - f_t(\x_t) \big) \\
& + \frac{2}{\lambda} \left( 1+ \frac{1}{\rho}\right)  \sum_{t=2}^T  \big( f_{t-1}(\u_{t-1}) - f_{t-1}(\x_{t-1}) \big)   \\
\leq & \sum_{t=1}^T f_t(\x_t)+\frac{1+\rho}{2}   \sum_{t=1}^T  \|\u_{t} -\u_{t-1}\|^2  + \frac{4}{\lambda} \left( 1+ \frac{1}{\rho}\right)  \sum_{t=1}^T  \big( f_t(\u_{t}) - f_t(\x_t) \big) \\
=&   \frac{4}{\lambda} \left( 1+ \frac{1}{\rho}\right) \sum_{t=1}^T  f_t(\u_{t}) + \frac{1+\rho}{2}   \sum_{t=1}^T  \|\u_{t} -\u_{t-1}\|^2 + \left( 1- \frac{4}{\lambda} \left( 1+ \frac{1}{\rho}\right) \right) \sum_{t=1}^T f_t(\x_t).
\end{split}
\end{equation}

First, we consider the case that
\begin{equation} \label{eqn:quadratic:rho}
 1- \frac{4}{\lambda} \left( 1+ \frac{1}{\rho}\right)  \leq 0  \Leftrightarrow    \frac{\lambda }{4} \leq 1 +\frac{1}{\rho}
\end{equation}
and have
\[
\begin{split}
&\sum_{t=1}^T  \left( f_t(\x_t) + \frac{1}{2} \|\x_t - \x_{t-1}\|^2 \right) \\
 \overset{(\ref{eqn:naive:sum:quadratic}), (\ref{eqn:quadratic:rho})}{\leq}    & \frac{4}{\lambda} \left( 1+ \frac{1}{\rho}\right) \sum_{t=1}^T  f_t(\u_{t}) + \frac{1+\rho}{2}   \sum_{t=1}^T  \|\u_{t} -\u_{t-1}\|^2\\
\leq & \max\left( \frac{4}{\lambda} \left(1+ \frac{1}{\rho} \right), 1 + \rho\right) \sum_{t=1}^T \left( f_t(\u_{t}) + \frac{1}{2} \|\u_{t} -\u_{t-1}\|^2 \right).
\end{split}
\]
To minimize the competitive ratio, we set
\[
\frac{4}{\lambda} \left(1+ \frac{1}{\rho} \right) = 1 + \rho \Rightarrow \rho=\frac{4}{\lambda}
\]
and obtain
\begin{equation} \label{eqn:ratio:quadratic:navie:1}
\begin{split}
\sum_{t=1}^T  \left( f_t(\x_t) + \frac{1}{2} \|\x_t - \x_{t-1}\|^2 \right)  \leq \left( 1 + \frac{4}{\lambda} \right) \sum_{t=1}^T \left( f_t(\u_{t}) + \frac{1}{2} \|\u_{t} -\u_{t-1}\|^2 \right).
\end{split}
\end{equation}

Next, we study the case that
\[
 1- \frac{4}{\lambda} \left( 1+ \frac{1}{\rho}\right)  \geq 0  \Leftrightarrow    \frac{\lambda }{4} \geq 1 +\frac{1}{\rho}
\]
which only happens when $\lambda > 4$. Then, we have
\[
 \sum_{t=1}^T  \left( f_t(\x_t) + \frac{1}{2} \|\x_t - \x_{t-1}\|^2 \right)  \overset{(\ref{eqn:naive}), (\ref{eqn:naive:sum:quadratic})}{\leq}     \sum_{t=1}^T  f_t(\u_{t}) + \frac{1+\rho}{2}   \sum_{t=1}^T  \|\u_{t} -\u_{t-1}\|^2.
\]
To minimize the competitive ratio, we set $\rho  = \frac{4}{\lambda-4}$, and obtain
\[
\sum_{t=1}^T  \left( f_t(\x_t) + \frac{1}{2} \|\x_t - \x_{t-1}\|^2 \right)  \leq  \frac{\lambda}{\lambda -4 }   \sum_{t=1}^T \left( f_t(\u_{t}) + \frac{1}{2}     \|\u_{t} -\u_{t-1}\|^2 \right)
\]
which is worse than (\ref{eqn:ratio:quadratic:navie:1}). So, we keep (\ref{eqn:ratio:quadratic:navie:1}) as the final result.

\subsection{Proof of Theorem~\ref{thm:greedy:convex:quadratic}}
Since $f_t(\cdot)$ is convex, the objective function of (\ref{eqn:2}) is $\gamma$-strongly convex. From the quadratic growth property of strongly convex functions \citep{COLT:Hazan:2011}, we have
\begin{equation}\label{eqn:opt:convex:quadratic}
f_t(\x_t) + \frac{\gamma}{2}\|\x_t-\x_{t-1}\|^2 + \frac{\gamma}{2} \|\u-\x_t\|^2 \leq f_t(\u) + \frac{\gamma}{2}\|\u-\x_{t-1}\|^2, \ \forall \u \in \X.
\end{equation}

Similar to previous studies \citep{bansal_et_al:LIPIcs}, the analysis uses an amortized local competitiveness argument, using the potential function $c \|\x_t -\u_t\|^2$. We proceed to bound $f_t(\x_t) + \frac{1}{2} \|\x_t-\x_{t-1}\|^2 + c \|\x_t -\u_t\|^2 - c\|\x_{t-1}-\u_{t-1}\|^2$, and have
\[
\begin{split}
&  f_t(\x_t) + \frac{1}{2}\|\x_t-\x_{t-1}\|^2 + c \|\x_t -\u_t\|^2 - c\|\x_{t-1}-\u_{t-1}\|^2\\
\overset{(\ref{eqn:inequality:squared})}{\leq}  &  f_t(\x_t) + \frac{1}{2}\|\x_t-\x_{t-1}\|^2+ c \big( 2 \|\x_t - \bv_t\|^2 + 2\|\bv_t-\u_t\|^2 \big)- c\|\x_{t-1}-\u_{t-1}\|^2\\
\overset{(\ref{eqn:quadratic:growth})}{\leq}  & \left(1+\frac{4c }{\lambda} \right) f_t(\x_t) + \frac{1}{2} \|\x_t-\x_{t-1}\|^2 + \frac{4c}{\lambda} f_t (\u_t) - c\|\x_{t-1}-\u_{t-1}\|^2\\
= & \left(1 +\frac{4c }{\lambda} \right) \left( f_t(\x_t) + \frac{\lambda}{2(\lambda + 4c) } \|\x_t-\x_{t-1}\|^2 \right) + \frac{4c}{\lambda} f_t (\u_t) - c\|\x_{t-1}-\u_{t-1}\|^2.
\end{split}
\]

Suppose
\begin{equation} \label{eqn:convex:quadratic:constraint:1}
\frac{\lambda}{\lambda + 4c} \leq \gamma,
\end{equation}
we have
\[
\begin{split}
&  f_t(\x_t) + \frac{1}{2}\|\x_t-\x_{t-1}\|^2 + c \|\x_t -\u_t\|^2 - c\|\x_{t-1}-\u_{t-1}\|^2\\
\leq & \left(1+\frac{4c }{\lambda} \right) \left( f_t(\x_t) + \frac{\gamma}{2} \|\x_t-\x_{t-1}\|^2 \right)  + \frac{4c}{\lambda} f_t (\u_t)  - c\|\x_{t-1}-\u_{t-1}\|^2\\
\overset{(\ref{eqn:opt:convex:quadratic})}{\leq} & \left(1+\frac{4c }{\lambda}\right) \left( f_t(\u_t) + \frac{\gamma}{2} \|\u_t-\x_{t-1}\|^2 -\frac{\gamma}{2} \|\u_t-\x_t\|^2 \right)  + \frac{4c}{\lambda} f_t (\u_t)  - c\|\x_{t-1}-\u_{t-1}\|^2\\
= & \left(1+\frac{8c }{\lambda}\right) f_t (\u_t) + \frac{\gamma(\lambda + 4c) }{2\lambda} \|\u_t-\x_{t-1}\|^2  -  \frac{\gamma(\lambda + 4c) }{2\lambda} \|\u_t-\x_t\|^2- c\|\x_{t-1}-\u_{t-1}\|^2.
\end{split}
\]

Summing over all the iterations and assuming $\x_{0}=\u_{0}$, we have
\[
\begin{split}
& \sum_{t=1}^T \left( f_t(\x_t) + \frac{1}{2}\|\x_t-\x_{t-1}\|^2 \right) + c \|\x_T -\u_T\|^2  \\
\leq & \left(1+\frac{8c }{\lambda}\right) \sum_{t=1}^T f_t (\u_t) + \frac{\gamma(\lambda + 4c) }{2\lambda} \sum_{t=1}^T \|\u_t-\x_{t-1}\|^2 \\
 &-  \frac{\gamma(\lambda + 4c) }{2\lambda} \sum_{t=1}^T\|\u_t-\x_t\|^2- c \sum_{t=1}^T\|\x_{t-1}-\u_{t-1}\|^2\\
\leq & \left(1+\frac{8c }{\lambda}\right) \sum_{t=1}^T f_t (\u_t) + \frac{\gamma(\lambda + 4c) }{2\lambda} \sum_{t=1}^T \|\u_t-\x_{t-1}\|^2 -  \left(\frac{\gamma(\lambda + 4c) }{2\lambda} +c \right)\sum_{t=1}^T \|\x_{t-1}-\u_{t-1}\|^2 \\
\overset{(\ref{eqn:inequality:squared})}{\leq}  & \left(1+\frac{8c }{\lambda}\right) \sum_{t=1}^T f_t (\u_t) + \frac{\gamma(\lambda + 4c) }{2\lambda} \sum_{t=1}^T \|\u_t-\x_{t-1}\|^2 \\
&-  \left(\frac{\gamma(\lambda + 4c) }{2\lambda} +c \right)\sum_{t=1}^T \left(\frac{1}{1+\rho}\|\x_{t-1}-\u_t\|^2 -\frac{1}{\rho}\|\u_t - \u_{t-1}\|^2\right) \\
\end{split}
\]
\[
\begin{split}
\leq & \left(1+\frac{8c }{\lambda}\right) \sum_{t=1}^T f_t (\u_t)  + \left(\frac{\gamma(\lambda + 4c) }{2\lambda} +c \right) \frac{1}{\rho} \sum_{t=1}^T \|\u_t - \u_{t-1}\|^2  \\
\leq &  \max\left(1 +\frac{8c}{\lambda}, \left(\frac{\gamma(\lambda + 4c) }{2\lambda} +c \right) \frac{2}{\rho} \right) \sum_{t=1}^T  \left(f_t (\u_t) + \frac{1}{2}\|\u_t - \u_{t-1}\|^2\right)
\end{split}
\]
where in the penultimate inequality  we assume
\begin{equation} \label{eqn:convex:quadratic:constraint:2}
\frac{\gamma(\lambda + 4c) }{2\lambda} \leq \left(\frac{\gamma(\lambda + 4c) }{2\lambda} +c \right) \frac{1}{1+\rho} \Leftrightarrow \frac{\gamma(\lambda + 4c) }{2\lambda} \leq \frac{c}{\rho}.
\end{equation}

Next, we minimize the competitive ratio under the constraints in (\ref{eqn:convex:quadratic:constraint:1}) and (\ref{eqn:convex:quadratic:constraint:2}), which can be summarized as
\[
\frac{\lambda}{\lambda + 4c} \leq \gamma \leq  \frac{\lambda}{\lambda + 4c}  \frac{2c}{\rho}.
\]
We first set $c=\frac{\rho}{2}$ and $\gamma=\frac{\lambda}{\lambda + 4c}$, and obtain
\[
\sum_{t=1}^T \left( f_t(\x_t) + \frac{1}{2}\|\x_t-\x_{t-1}\|^2 \right)
 \leq  \max\left(1 +\frac{4 \rho}{\lambda},  1 + \frac{1 }{\rho} \right) \sum_{t=1}^T  \left(f_t (\u_t) + \frac{1}{2}\|\u_t - \u_{t-1}\|^2\right).
\]
Then, we set
\[
1 +\frac{4 \rho}{\lambda}= 1 + \frac{1 }{\rho} \Rightarrow  \rho= \frac{\sqrt{\lambda}}{2}.
\]
As a result, the competitive ratio is
\[
1 + \frac{1 }{\rho}  = 1 + \frac{2}{\sqrt{\lambda}},
\]
and the parameter is
\[
 \gamma=\frac{\lambda}{\lambda + 4c}  = \frac{\lambda}{\lambda + 2 \rho} = \frac{\lambda}{\lambda + \sqrt{\lambda}}  .
\]

\subsection{Proof of Theorem~\ref{thm:ader}}
The analysis is similar to the proof of Theorem 3 of \citet{Adaptive:Dynamic:Regret:NIPS}. In the analysis, we need to specify the behavior of the meta-algorithm and expert-algorithm at $t=0$. To simplify the presentation, we set
\begin{equation} \label{eqn:condit:at:zero}
\x_0=0, \textrm{ and } \x_0^\eta=0, \ \forall \eta \in \H.
\end{equation}

First, we bound the dynamic regret with switching cost of the meta-algorithm w.r.t.~all experts simultaneously.
\begin{lemma} \label{lem:meta} Under Assumptions~\ref{ass:1} and \ref{ass:2}, and setting $\beta= \frac{2}{(2G+1) D} \sqrt{\frac{2 }{5 T}}$, we have
\begin{equation}\label{thm:main:1}
 \sum_{t=1}^T \Big( s_t(\x_t) + \|\x_t -\x_{t-1}\|\Big) - \sum_{t=1}^T \Big(s_t(\x_t^\eta)  + \|\x_t^\eta- \x_{t-1}^\eta\| \Big) \leq  (2G+1) D\sqrt{\frac{5  T }{8}} \left( \ln \frac{1}{w_1^\eta} + 1\right)
\end{equation}
for each $\eta \in \H$.
\end{lemma}

Next, we bound the dynamic regret with switching cost of each expert w.r.t.~any comparator sequence $\u_0, \u_1,\ldots,\u_T \in \X$.
\begin{lemma} \label{lem:expert} Under Assumptions \ref{ass:1} and \ref{ass:2},  we have
\begin{equation}\label{eqn:main:2}
\sum_{t=1}^T \Big(s_t(\x_t^\eta)  + \|\x_t^\eta- \x_{t-1}^\eta\| \Big) - \sum_{t=1}^T  s_t(\u_t) \leq \frac{D^2}{2 \eta}    + \frac{D}{\eta} \sum_{t=1}^T  \|\u_t - \u_{t-1}\| + \eta T \left( \frac{G^2}{2}  +G\right).
\end{equation}
\end{lemma}

Then, we show that for any sequence of  comparators $\u_0, \u_1,\ldots,\u_T \in \X$ there exists an $\eta_k \in \H$ such that the R.H.S.~of (\ref{eqn:main:2}) is almost minimal. If we minimize the R.H.S.~of (\ref{eqn:main:2}) exactly, the optimal step size is
\begin{equation}\label{thm:main:3}
\eta^*(P_T)= \sqrt{\frac{D^2+ 2 D P_T}{T (G^2+2G)}}.
\end{equation}
From Assumption~\ref{ass:2}, we have the following bound of the path-length
\begin{equation}\label{thm:main:9}
0 \leq P_T= \sum_{t=1}^T  \|\u_t - \u_{t-1}\| \overset{(\ref{eqn:domain})}{\leq} TD.
\end{equation}
Thus
\[
\sqrt{\frac{D^2}{T (G^2+2G)}}\leq \eta^*(P_T) \leq \sqrt{\frac{D^2+2TD^2}{T (G^2+2G)}}.
\]
From our construction of $\H$ in (\ref{eqn:def:H}), it is easy to verify that
\[
\min \H= \sqrt{\frac{D^2}{T (G^2+2G)}}, \textrm{ and  } \max \H \geq \sqrt{\frac{D^2+2TD^2}{T (G^2+2G)}}.
\]
As a result, for any possible value of $P_T$, there exists a step size $\eta_k \in \H$ with $k$ defined in (\ref{eqn:def:k}), such that
\begin{equation}\label{thm:main:4}
\eta_k = 2^{k-1}  \sqrt{\frac{D^2}{T (G^2+2G)}}  \leq \eta^*(P_T) \leq 2 \eta_k.
\end{equation}

Plugging $\eta_k$ into (\ref{eqn:main:2}), the dynamic regret with switching cost of expert $E^{\eta_k}$ is given by
\begin{equation}\label{thm:main:5}
\begin{split}
& \sum_{t=1}^T \Big(s_t(\x_t^{\eta_k})  + \|\x_t^{\eta_k}- \x_{t-1}^{\eta_k}\| \Big) - \sum_{t=1}^T  s_t(\u_t) \\
\leq &\frac{D^2}{2 \eta_k}    + \frac{D}{{\eta_k}} \sum_{t=1}^T  \|\u_t - \u_{t-1}\| + \eta_k T \left( \frac{G^2}{2}  +G\right)\\
\overset{(\ref{thm:main:4})}{\leq} & \frac{D^2}{\eta^*(P_T)}    + \frac{2D}{{\eta^*(P_T)}} \sum_{t=1}^T  \|\u_t - \u_{t-1}\| + \eta^*(P_T) T \left( \frac{G^2}{2}  +G\right) \\
 \overset{(\ref{thm:main:3})}{=} & \frac{3}{2}\sqrt{T (G^2+2G)(D^2+ 2 D P_T)}.
\end{split}
\end{equation}

From (\ref{eqn:initial:weight}), we know the initial weight of expert $E^{\eta_k}$ is
\[
w_1^{\eta_k}=\frac{C}{k(k+1)} \geq \frac{1}{k(k+1)} \geq \frac{1}{(k+1)^2}.
\]
Combining with (\ref{thm:main:1}), we obtain the relative performance of the meta-algorithm w.r.t.~expert $E^{\eta_k}$:
\begin{equation}\label{thm:main:6}
 \sum_{t=1}^T \Big( s_t(\x_t) + \|\x_t -\x_{t-1}\|\Big) - \sum_{t=1}^T \Big(s_t(\x_t^{\eta_k})  + \|\x_t^{\eta_k}- \x_{t-1}^{\eta_k}\| \Big) \leq  (2G+1) D\sqrt{\frac{5  T }{8}} \left[1+ 2\ln (k+1)\right].
\end{equation}

From  (\ref{thm:main:5}) and (\ref{thm:main:6}), we derive the following upper bound for dynamic regret with switching cost
\begin{equation}\label{thm:main:7}
\begin{split}
& \sum_{t=1}^T \Big( s_t(\x_t) + \|\x_t -\x_{t-1}\|\Big)-  \sum_{t=1}^T  s_t(\u_t) \\
\leq & \frac{3}{2}\sqrt{T (G^2+2G)(D^2+ 2 D P_T)}+ (2G+1) D\sqrt{\frac{5  T }{8}} \left[1+ 2\ln (k+1)\right].
\end{split}
\end{equation}
Finally, from Assumption~\ref{ass:0}, we have
\begin{equation}\label{thm:main:8}
f_t(\x_t) - f_t(\u_t) \leq \langle \nabla f_t(\x_t), \x_t-\u_t \rangle  \overset{(\ref{eqn:surrogate})}{=}   s_t(\x_t) - s_t(\u_t).
\end{equation}
We complete the proof by combining (\ref{thm:main:7}) and (\ref{thm:main:8}).

\subsection{Proof of Theorem~\ref{thm:lookahead:ader}}
The analysis is similar to that of Theorem~\ref{thm:ader}. The difference is that we need to take into account the lookahead property of the meta-algorithm and the expert-algorithm.

First, we bound the dynamic regret with switching cost of the meta-algorithm w.r.t.~all experts simultaneously.
\begin{lemma} \label{lem:lookhead:meta} Under Assumption~\ref{ass:2}, and setting $\beta= \frac{1}{D}  \sqrt{\frac{2}{T}}$, we have
\begin{equation}\label{eqn:lookhead:1}
 \sum_{t=1}^T \Big( s_t(\x_t) + \|\x_t -\x_{t-1}\|\Big) - \sum_{t=1}^T \Big(s_t(\x_t^\eta)  + \|\x_t^\eta- \x_{t-1}^\eta\| \Big)  \leq D \sqrt{ \frac{T}{2}}  \left( \ln \frac{1}{w_0^\eta} +1 \right)
\end{equation}
for each $\eta \in \H$.
\end{lemma}
Combining Lemma~\ref{lem:lookhead:meta} with Assumption~\ref{ass:0}, we have
\begin{equation}\label{eqn:lookhead:2}
 \sum_{t=1}^T \Big( f_t(\x_t) + \|\x_t -\x_{t-1}\|\Big) - \sum_{t=1}^T \Big(f_t(\x_t^\eta)  + \|\x_t^\eta- \x_{t-1}^\eta\| \Big) \overset{(\ref{thm:main:8}), (\ref{eqn:lookhead:1})}{\leq}   D \sqrt{ \frac{T}{2}}  \left( \ln \frac{1}{w_0^\eta} +1 \right)
\end{equation}
for each $\eta \in \H$.

Next, we bound the dynamic regret with switching cost of each expert w.r.t.~any comparator sequence $\u_0, \u_1,\ldots,\u_T \in \X$.
\begin{lemma} \label{lem:lookhead:expert} Under Assumptions~\ref{ass:0} and \ref{ass:2},  we have
\begin{equation}\label{eqn:lookhead:3}
\sum_{t=1}^T \Big(f_t(\x_t^\eta)  + \|\x_t^\eta- \x_{t-1}^\eta\| \Big) - \sum_{t=1}^T  f_t(\u_t) \leq \frac{D^2}{2 \eta}    + \frac{D}{\eta} \sum_{t=1}^T \|\u_t - \u_{t-1}\|  + \frac{\eta T}{2}.
\end{equation}
\end{lemma}

The rest of the proof is almost identical to that of Theorem~\ref{thm:ader}. We will show that  for any sequence of  comparators $\u_0, \u_1,\ldots,\u_T \in \X$ there exists an $\eta_k \in \H$ such that the R.H.S.~of (\ref{eqn:lookhead:3}) is almost minimal. If we minimize the R.H.S.~of (\ref{eqn:lookhead:3}) exactly, the optimal step size is
\begin{equation}\label{eqn:lookhead:4}
\eta^*(P_T)= \sqrt{\frac{D^2+ 2 D P_T}{T}}.
\end{equation}
From (\ref{thm:main:9}), we know that
\[
\sqrt{\frac{D^2}{T}}\leq \eta^*(P_T) \leq \sqrt{\frac{D^2+2TD^2}{T}}.
\]
From our construction of $\H$ in (\ref{eqn:def:H:New}), it is easy to verify that
\[
\min \H= \sqrt{\frac{D^2}{T }}, \textrm{ and  } \max \H \geq \sqrt{\frac{D^2+2TD^2}{T}}.
\]
As a result, for any possible value of $P_T$, there exists a step size $\eta_k \in \H$ with $k$ defined in (\ref{eqn:def:k}), such that
\begin{equation}\label{eqn:lookhead:5}
\eta_k = 2^{k-1}  \sqrt{\frac{D^2}{T}}  \leq \eta^*(P_T) \leq 2 \eta_k.
\end{equation}

Plugging $\eta_k$ into (\ref{eqn:lookhead:3}), the dynamic regret with switching cost of expert $E^{\eta_k}$ is given by
\begin{equation}\label{eqn:lookhead:6}
\begin{split}
& \sum_{t=1}^T \Big(f_t(\x_t^{\eta_k})  + \|\x_t^{\eta_k}- \x_{t-1}^{\eta_k}\| \Big) - \sum_{t=1}^T  f_t(\u_t) \\
\leq &\frac{D^2}{2 \eta_k}    + \frac{D}{{\eta_k}} \sum_{t=1}^T  \|\u_t - \u_{t-1}\| + \frac{\eta_k T}{2}\\
\overset{(\ref{eqn:lookhead:5})}{\leq} & \frac{D^2}{\eta^*(P_T)} + \frac{2D}{{\eta^*(P_T)}} \sum_{t=1}^T  \|\u_t - \u_{t-1}\| + \frac{\eta^*(P_T) T}{2}\\
 \overset{(\ref{eqn:lookhead:4})}{=} & \frac{3}{2}\sqrt{T (D^2+ 2 D P_T)}.
\end{split}
\end{equation}

From Step 2 of Algorithm~\ref{alg:3}, we know the initial weight of expert $E^{\eta_k}$ is
\[
w_0^{\eta_k}=\frac{C}{k(k+1)} \geq \frac{1}{k(k+1)} \geq \frac{1}{(k+1)^2}.
\]
Combining with (\ref{eqn:lookhead:2}), we obtain the relative performance of the meta-algorithm w.r.t.~expert $E^{\eta_k}$:
\begin{equation}\label{eqn:lookhead:7}
 \sum_{t=1}^T \Big( f_t(\x_t) + \|\x_t -\x_{t-1}\|\Big) - \sum_{t=1}^T \Big(f_t(\x_t^{\eta_k})  + \|\x_t^{\eta_k}- \x_{t-1}^{\eta_k}\| \Big) \leq D \sqrt{ \frac{T}{2}}  \left[1+ 2\ln (k+1)\right].
\end{equation}
We complete the proof by summing (\ref{eqn:lookhead:6}) and (\ref{eqn:lookhead:7}) together.

\subsection{Proof of Theorem~\ref{thm:lower}}
The proof is built upon a lower bound of competitive ratio \citep{pmlr-v125-argue20a}. By setting $\gamma=\frac{D}{2 \sqrt{d}}$ in Lemma 12 of \citet{pmlr-v125-argue20a}, we can guarantee that Assumption~\ref{ass:2} is satisfied.  Then, we choose $\mu=0$, $\lambda=1/\gamma$ in that lemma, and obtain the conclusion below.

\begin{lemma} \label{thm:lower:ratio} For any online algorithm $A$ and any fixed value of $d$, there exists a sequence of convex functions $f_1(\cdot), \ldots, f_d(\cdot)$ over the domain $[-\frac{D}{2 \sqrt{d}},\frac{D}{2 \sqrt{d}}]^d$ in the lookahead setting such that
\begin{compactenum}
  \item the sum of the hitting cost and the switching cost of $A$ is at least $\frac{3 \gamma d}{4}=\frac{3 D \sqrt{d}}{8}$;
  \item there exist a fixed point $\u$ whose hitting cost is $0$.
\end{compactenum}
\end{lemma}

We consider two cases: $\tau < D$ and $\tau \geq D$. When $\tau < D$, from Lemma~\ref{thm:lower:ratio} with $d=T$, we know that the dynamic regret with switching cost w.r.t.~a fixed point $\u$ is at least $\Omega(D \sqrt{T})$.

Next, we consider the case $\tau \geq D$.   Without loss of generality, we assume $\lfloor \tau/D \rfloor$ divides $T$. Then, we partition $T$ into $\lfloor\tau/D \rfloor$ successive stages, each of which contains $T/\lfloor \tau/D \rfloor$ rounds. Applying Lemma~\ref{thm:lower:ratio} to each stage, we conclude that there exists a sequence of convex functions $f_1(\cdot), \ldots, f_T(\cdot)$ over the domain $[-\frac{D}{2 \sqrt{d}},\frac{D}{2 \sqrt{d}}]^d$ where $d=T/\lfloor \tau/D \rfloor$ in the lookahead setting such that
 \begin{compactenum}
  \item the sum of the hitting cost and the switching cost of any online algorithm is at least
\[
\left \lfloor \tau/D \right \rfloor \cdot  \frac{3 D }{8} \sqrt{\frac{T}{\lfloor \tau/D \rfloor}} =  \frac{3 D }{8} \sqrt{T \left \lfloor \frac{\tau}{D} \right \rfloor } =\Omega(\sqrt{TD\tau});
\]
  \item there exists a sequence of points $\u_1,\ldots,\u_T$ whose hitting cost is $0$ and switching cost (i.e., path-length) is at most
\[
D \left \lfloor \frac{\tau}{D} \right \rfloor  \leq \tau
\]
since they switch at most $\lfloor\tau/D \rfloor-1$ times.
\end{compactenum}
Thus, the dynamic regret with switching cost w.r.t.~$\u_1,\ldots,\u_T$ is at least
\[
 \frac{3 D }{8} \sqrt{T \left \lfloor \frac{\tau}{D} \right \rfloor } - \tau =\Omega(\sqrt{TD\tau}).
\]

We complete the proof by combining the results of the above two cases.
\section{Conclusion and Future Work}
We investigate the problem of smoothed online learning (SOL), and derive constant competitive ratio or sublinear dynamic regret with switching cost. For competitive ratio, we demonstrate that the naive approach, which only minimizes the hitting cost, is $\max(1, \frac{2}{\alpha})$-competitive for $\alpha$-polyhedral functions and $1 + \frac{4}{\lambda}$-competitive for $\lambda$-quadratic growth functions. Furthermore, we show that the greedy algorithm, which minimizes the weighted sum of the hitting cost and the switching cost, is $1+\frac{2}{\sqrt{\lambda}}$-competitive for convex and $\lambda$-quadratic growth functions. For dynamic regret with switching cost, we propose smoothed Ader (SAder), which attains the optimal $O(\sqrt{T(1+P_T)})$ bound. We also develop a lookahead version of SAder to make use of the prior knowledge of the hitting cost, and establish an $\Omega(\sqrt{T(1+P_T)})$ lower bound. All the proposed algorithms are very simple, and our results served as the baselines for SOL.

The research on SOL is still on its early stage, and there are many open problems.
\begin{compactenum}
  \item Although we can upper bound the sum of the hitting cost and the switching cost, we do not have a direct control over the switching cost. However, in many real problems, there may exist a hard constraint on the switching cost, motivating the study of switch-constrained online learning, in which the times of switches is limited \citep{Online:Limited:Switching,NEURIPS2020_236f119f}. It would be interesting to investigate how to impose a budget on the switching cost.
  \item The current analysis of competitive ratio and dynamic regret with switching cost relies on very different conditions. Generally speaking, to bound the competitive ratio, we need stronger assumptions, such as polyhedrality  or strong convexity. It is unclear how to make use of those conditions when bounding the dynamic regret.
  \item As aforementioned, for polyhedral functions and quadratic growth functions, the best competitive ratio is obtained by the naive approach which ignores the switching cost, which is counterintuitive. To better understand the challenge, it is important to reveal the lower bound of polyhedral functions and quadratic growth functions.
  \item We note that SOL has a similar spirit with continual learning \citep{Kirkpatrick3521,NEURIPS2019_e562cd9c}, which aims to learn consecutive tasks (corresponding to minimize the hitting cost) without forgetting how to perform previously trained tasks (corresponding to minimize the switching cost). So, the theoretical studies of SOL may lay the foundation for continual learning.
\end{compactenum}

\acks{The authors would like to thank Yuxuan Xiang for discussions about Theorem~\ref{thm:ader}.}
\bibliography{E:/MyPaper/ref}
\appendix
\section{Proof of Lemma~\ref{lem:meta}}
Based on the prediction rule of the meta-algorithm, we upper bound the switching cost when $t\geq 2$ as follows:
\begin{equation} \label{eqn:decompose:switching}
\begin{split}
& \|\x_t -\x_{t-1}\| =\left\| \sum_{\eta \in \H} w_t^\eta \x_t^\eta- \sum_{\eta \in \H} w_{t-1}^\eta \x_{t-1}^\eta \right\|  = \left\| \sum_{\eta \in \H} w_t^\eta (\x_t^\eta-\x)- \sum_{\eta \in \H} w_{t-1}^\eta (\x_{t-1}^\eta -\x) \right\|\\
\leq & \left\| \sum_{\eta \in \H} w_t^\eta (\x_t^\eta - \x)- \sum_{\eta \in \H} w_t^\eta (\x_{t-1}^\eta- \x) \right\| + \left\| \sum_{\eta \in \H} w_t^\eta (\x_{t-1}^\eta- \x) - \sum_{\eta \in \H} w_{t-1}^\eta (\x_{t-1}^\eta- \x) \right\| \\
=& \left\| \sum_{\eta \in \H} w_t^\eta (\x_t^\eta -   \x_{t-1}^\eta) \right\| + \left\| \sum_{\eta \in \H} (w_t^\eta -w_{t-1}^\eta)(\x_{t-1}^\eta- \x) \right\| \\
\leq &  \sum_{\eta \in \H} w_t^\eta \left\| \x_t^\eta -   \x_{t-1}^\eta \right\| + \sum_{\eta \in \H} |w_t^\eta -w_{t-1}^\eta|\left\| \x_{t-1}^\eta- \x  \right\| \\
\overset{(\ref{eqn:domain})}{\leq} &  \sum_{\eta \in \H} w_t^\eta \left\| \x_t^\eta -   \x_{t-1}^\eta \right\| + D \sum_{\eta \in \H} |w_t^\eta -w_{t-1}^\eta|= \sum_{\eta \in \H} w_t^\eta \left\| \x_t^\eta -   \x_{t-1}^\eta \right\| + D \|\w_t -\w_{t-1}\|_1
\end{split} \!\!\!\!\!\!\!\!\!\!\!
\end{equation}
where $\x$ is an arbitrary point in $\X$, and $\w_t=(w_t^{\eta})_{\eta \in \H} \in \R^N$. When $t=1$, from (\ref{eqn:condit:at:zero}), we have
\begin{equation} \label{eqn:decompose:switching:t1}
\|\x_1 -\x_{0}\| = \|\x_1\| = \left \| \sum_{\eta \in \H} w_1^\eta \x_1^\eta \right \| \leq \sum_{\eta \in \H} w_1^\eta \left\| \x_1^\eta \right\|=\sum_{\eta \in \H} w_1^\eta \left\| \x_1^\eta - \x_0^\eta \right\| .
\end{equation}

Then, the relative loss of the meta-algorithm w.r.t.~expert $E^\eta$ can be decomposed as
\begin{equation} \label{eqn:meta:loss}
\begin{split}
& \sum_{t=1}^T \Big( s_t(\x_t) + \|\x_t -\x_{t-1}\|\Big) - \sum_{t=1}^T \Big(s_t(\x_t^\eta)  + \|\x_t^\eta- \x_{t-1}^\eta\| \Big) \\
\overset{(\ref{eqn:surrogate}), (\ref{eqn:decompose:switching}),(\ref{eqn:decompose:switching:t1})}{\leq} &  \sum_{t=1}^T \left(\sum_{\eta \in \H} w_t^\eta  \left\| \x_t^\eta -   \x_{t-1}^\eta \right\|  - \big( \langle \nabla f_t(\x_t), \x_t^\eta -\x_t \rangle +  \|\x_t^\eta- \x_{t-1}^\eta\| \big) \right) \\
&+  D \sum_{t=2}^T  \|\w_t -\w_{t-1}\|_1 \\
\overset{(\ref{eqn:meta:loss:function})}{=} &  \underbrace{\sum_{t=1}^T \left(\sum_{\eta \in \H} w_t^\eta  \ell_t(\x_t^\eta)  - \ell_t(\x_t^\eta)  \right)}_{:=A}  +  D \sum_{t=2}^T  \|\w_t -\w_{t-1}\|_1.
\end{split}
\end{equation}

We proceed to bound $A$ and $\|\w_t -\w_{t-1}\|_1$ in (\ref{eqn:meta:loss}). Notice that $A$ is the regret of the meta-algorithm w.r.t.~expert $E^\eta$. From Assumptions~\ref{ass:1} and \ref{ass:2}, we have
\[
\left| \langle \nabla f_t(\x_t), \x_t^\eta -\x_t \rangle  \right| \leq \| \nabla f_t(\x_t)\| \|\x_t^\eta -\x_t\|  \overset{(\ref{eqn:grad}), (\ref{eqn:domain})}{\leq} GD.
\]
Thus, we have
\begin{equation} \label{eqn:bound:loss}
-GD \leq \ell_t(\x_t^\eta)  \leq (G+1)D, \ \forall \eta \in \H.
\end{equation}
According to the standard analysis of Hedge \citep[Lemma 1]{Adaptive:Dynamic:Regret:NIPS} and (\ref{eqn:bound:loss}), we have
\begin{equation} \label{eqn:bound:A}
\sum_{t=1}^T \left(\sum_{\eta \in \H} w_t^\eta  \ell_t(\x_t^\eta)  - \ell_t(\x_t^\eta)  \right) \leq \frac{1}{\beta} \ln \frac{1}{w_1^\eta}  + \frac{\beta T (2G+1)^2 D^2}{8}.
\end{equation}

Next, we bound $\|\w_t -\w_{t-1}\|_1$, which measures the stability of the meta-algorithm, i.e., the change of coefficients between successive rounds.  Because the Hedge algorithm is translation invariant, we can subtract $D/2$ from $\ell_t(\x_t^\eta)$ such that
\begin{equation} \label{eqn:bound:loss2}
\left| \ell_t(\x_t^\eta) - D/2 \right| \leq (G+1/2)D, \ \forall \eta \in \H.
\end{equation}
It is well-known that Hedge can be treated as a special case of ``Follow-the-Regularized-Leader'' with  entropic regularization \citep{Online:suvery}
\[
R(\w)=\sum_i w_i \log w_i
\]
over the probability simplex, and $R(\cdot)$ is $1$-strongly convex  w.r.t.~the $\ell_1$-norm. In other words, we have
 \[
 \w_{t+1}= \argmin_{\w \in \Delta} \left \langle - \frac{1}{\beta} \log (\w_1) + \sum_{i=1}^t \g_i , \w \right \rangle + \frac{1}{\beta} R(\w), \ \forall t \geq 1
 \]
 where $\Delta \subseteq \R^{N}$ is the probability simplex, and $\g_i=[\ell_i(\x_i^\eta)- D/2]_{\eta \in \H} \in \R^N$.  From the stability property of Follow-the-Regularized-Leader \citep[Lemma 2]{DV:Distributed}, we have
\[
\|\w_t -\w_{t-1}\|_1 \leq \beta \|\g_{t-1}\|_{\infty} \overset{(\ref{eqn:bound:loss2})}{\leq} \beta (G+1/2)D, \ \forall t \geq 2.
\]
Then
\begin{equation}\label{eqn:bound:sum:b}
 \sum_{t=2}^T \|\w_t -\w_{t-1}\|_1 \leq  \frac{ \beta (T-1)(2G +1)D}{2}.
\end{equation}

Substituting  (\ref{eqn:bound:A}) and (\ref{eqn:bound:sum:b}) into (\ref{eqn:meta:loss}), we have
\[
\begin{split}
& \sum_{t=1}^T \Big( s_t(\x_t) + \|\x_t -\x_{t-1}\|\Big) - \sum_{t=1}^T \Big(s_t(\x_t^\eta)  + \|\x_t^\eta- \x_{t-1}^\eta\| \Big)\\
\leq  &  \frac{1}{\beta} \ln \frac{1}{w_1^\eta} + \frac{\beta T (2G+1)^2 D^2}{8} + \frac{ \beta (T-1)(2G +1)D^2}{2}   \leq \frac{1}{\beta} \ln \frac{1}{w_1^\eta} + \frac{5 \beta T (2G+1)^2 D^2}{8}.
\end{split}
\]
We complete the proof by setting $\beta= \frac{2}{(2G+1) D} \sqrt{\frac{2 }{5 T}}$.
\section{Proof of Lemma~\ref{lem:expert}}
First, we bound the dynamic regret of the expert-algorithm. Define
\[
\xb_{t+1}^\eta=\x_t^\eta - \eta \nabla f_t(\x_t).
\]
Following the analysis of Ader \citep[Theorems 1 and 6]{Adaptive:Dynamic:Regret:NIPS}, we have
\[
\begin{split}
& s_t(\x_t^\eta) - s_t(\u_t)\overset{(\ref{eqn:surrogate})}{=} \langle \nabla f_t(\x_t), \x_t^\eta - \u_t\rangle = \frac{1}{\eta} \langle \x_t^\eta  - \xb_{t+1}^\eta, \x_t^\eta - \u_t\rangle \\
= & \frac{1}{2 \eta} \left( \|\x_t^\eta-\u_t\|_2^2 - \|\xb_{t+1}^\eta-\u_t\|_2^2 + \|\x_t^\eta  - \xb_{t+1}^\eta\|_2^2 \right)\\
 =& \frac{1}{2 \eta} \left( \|\x_t^\eta-\u_t\|_2^2 - \|\xb_{t+1}^\eta-\u_t\|_2^2 \right) + \frac{\eta}{2 } \|\nabla f_t(\x_t)\|_2^2  \\
\overset{(\ref{eqn:grad})}{\leq} & \frac{1}{2 \eta} \left( \|\x_t^\eta-\u_t\|_2^2 - \|\xb_{t+1}^\eta-\u_t\|_2^2 \right) + \frac{\eta}{2 } G^2  \\
\leq & \frac{1}{2 \eta} \left( \|\x_t^\eta-\u_t\|_2^2 - \|\x_{t+1}^\eta-\u_t\|_2^2 \right) + \frac{\eta}{2 } G^2 \\
= & \frac{1}{2 \eta} \left( \|\x_t^\eta-\u_t\|_2^2 - \|\x_{t+1}^\eta-\u_{t+1}\|_2^2 + \|\x_{t+1}^\eta-\u_{t+1}\|_2^2 -\|\x_{t+1}^\eta-\u_t\|_2^2 \right) + \frac{\eta}{2 } G^2 \\
= & \frac{1}{2 \eta} \left( \|\x_t^\eta-\u_t\|_2^2 - \|\x_{t+1}^\eta-\u_{t+1}\|_2^2 + (\x_{t+1}^\eta-\u_{t+1} +\x_{t+1}^\eta-\u_t )^\top (\u_t-\u_{t+1})  \right) + \frac{\eta}{2 } G^2 \\
\leq  & \frac{1}{2 \eta} \left( \|\x_t^\eta-\u_t\|_2^2 - \|\x_{t+1}^\eta-\u_{t+1}\|_2^2 + \big(\|\x_{t+1}^\eta-\u_{t+1}\| + \|\x_{t+1}^\eta-\u_t \|\big) \|\u_t-\u_{t+1}\| \right) + \frac{\eta}{2 } G^2 \\
\overset{(\ref{eqn:domain})}{\leq}  & \frac{1}{2 \eta} \left( \|\x_t^\eta-\u_t\|_2^2 - \|\x_{t+1}^\eta-\u_{t+1}\|_2^2 \right) + \frac{D}{ \eta} \|\u_t-\u_{t+1}\| + \frac{\eta}{2 } G^2 .
\end{split}
\]
Summing the above inequality over all iterations, we have
\begin{equation} \label{eqn:expert:1}
\begin{split}
\sum_{t=1}^T \left( s_t(\x_t^\eta) - s_t(\u_t) \right) \leq & \frac{1}{2 \eta}   \|\x_1^\eta-\u_1\|_2^2  + \frac{D}{ \eta} \sum_{t=1}^T \|\u_{t+1} - \u_t\| + \frac{\eta T}{2 } G^2 \\
\overset{(\ref{eqn:domain})}{\leq} &  \frac{1}{2 \eta} D^2    + \frac{D}{ \eta} \sum_{t=1}^T \|\u_{t+1} - \u_t\| + \frac{\eta T}{2 } G^2.
\end{split}
\end{equation}
Since  (\ref{eqn:expert:1})  holds when $\u_{T+1} =\u_T$, we have
\begin{equation} \label{eqn:expert:2}
\begin{split}
\sum_{t=1}^T \left( s_t(\x_t^\eta) - s_t(\u_t) \right) \leq  \frac{1}{2 \eta} D^2    + \frac{D}{ \eta} \sum_{t=1}^T \|\u_t - \u_{t-1}\| + \frac{\eta T}{2 } G^2.
\end{split}
\end{equation}

Next, we bound the switching cost of the expert-algorithm. To this end, we have
\begin{equation} \label{eqn:expert:3}
\begin{split}
& \sum_{t=1}^T \|\x_t^\eta- \x_{t-1}^\eta\|  = \sum_{t=0}^{T-1} \|\x_{t+1}^\eta-\x_{t}^\eta\| \leq   \sum_{t=0}^{T-1} \|\xb_{t+1}^\eta-\x_{t}^\eta\| =  \sum_{t=0}^{T-1} \|\eta \nabla f_t(\x_t)\| \overset{(\ref{eqn:grad})}{\leq} \eta T G.
\end{split}
\end{equation}
We complete the proof by combining (\ref{eqn:expert:2}) with (\ref{eqn:expert:3}).
\section{Proof of Lemma~\ref{lem:lookhead:meta}}
We reuse the first part of the proof of Lemma~\ref{lem:meta}, and start from (\ref{eqn:meta:loss}). To bound $A$, we need to analyze the behavior of the lookahead Hedge. To this end, we prove the following lemma.

\begin{lemma}\label{lem:lookhead:hedge} The meta-algorithm in Algorithm~\ref{alg:3} satisfies
\begin{equation} \label{eqn:lookhead:hedge}
\sum_{t=1}^T \left(\sum_{\eta \in \H} w_t^\eta  \ell_t(\x_t^\eta)  - \ell_t(\x_t^\eta)  \right)  \leq  \frac{1}{\beta} \ln \frac{1}{w_0^\eta} - \frac{1}{2\beta} \sum_{t=1}^T \|\w_t-\w_{t-1}\|_1^2
\end{equation}
for any $\eta \in \H$.
\end{lemma}

Substituting (\ref{eqn:lookhead:hedge}) into (\ref{eqn:meta:loss}), we have
\begin{equation} \label{eqn:meta:loss:new}
\begin{split}
& \sum_{t=1}^T \Big( s_t(\x_t) + \|\x_t -\x_{t-1}\|\Big) - \sum_{t=1}^T \Big(s_t(\x_t^\eta)  + \|\x_t^\eta- \x_{t-1}^\eta\| \Big) \\
\leq & \frac{1}{\beta} \ln \frac{1}{w_0^\eta} - \frac{1}{2\beta} \sum_{t=1}^T \|\w_t-\w_{t-1}\|_1^2  +  D \sum_{t=2}^T  \|\w_t -\w_{t-1}\|_1 \\
\leq & \frac{1}{\beta} \ln \frac{1}{w_0^\eta} - \frac{1}{2\beta} \sum_{t=1}^T \|\w_t-\w_{t-1}\|_1^2  +   \sum_{t=2}^T \left( \frac{1}{2\beta} \|\w_t -\w_{t-1}\|_1^2 + \frac{\beta D^2}{2} \right) \\
\leq &  \frac{1}{\beta} \ln \frac{1}{w_0^\eta} + \frac{\beta T D^2}{2} =D \sqrt{ \frac{T}{2}}  \left( \ln \frac{1}{w_0^\eta} +1 \right)
\end{split}
\end{equation}
where we set $\beta= \frac{1}{D}  \sqrt{\frac{2}{T}}$.

\section{Proof of Lemma~\ref{lem:lookhead:hedge}}
To simplify the notation, we define
\[
W_0= \sum_{\eta \in \H} w_0^\eta =1, \ L_t^\eta= \sum_{i=1}^t \ell_i(\x_i^\eta), \textrm{ and } W_t= \sum_{\eta \in \H} w_0^\eta e^{- \beta L_t^\eta}, \ \forall t \geq 1.
\]
From the updating rule in (\ref{eqn:lookahead:hedge:pro}), it is easy to verify that
\begin{equation} \label{eqn:wt:eta}
w_t^\eta =  \frac{w_0^\eta  e^{-\beta L_{t}^\eta}}{W_t}, \ \forall t \geq 1.
\end{equation}

First, we have
\begin{equation} \label{eqn:wt:w0}
\begin{split}
\ln W_T = \ln \left( \sum_{\eta \in \H} w_0^\eta e^{- \beta L_T^\eta}\right) \geq \ln\left( \max_{\eta \in \H}  w_0^\eta e^{- \beta L_T^\eta} \right) =- \beta \min_{\eta \in \H} \left(L_T^\eta + \frac{1}{\beta} \ln \frac{1}{w_0^\eta} \right) .
\end{split}
\end{equation}

Next, we bound the related quantity $\ln(W_t/W_{t-1})$ as follows.  For any $\eta \in \H$, we have
\begin{equation} \label{eqn:wt:wt1}
\ln \left(\frac{W_t}{W_{t-1}} \right)\overset{(\ref{eqn:wt:eta})}{=} \ln \left(\frac{w_0^\eta  e^{-\beta L_{t}^\eta}}{w_t^\eta}  \frac{w_{t-1}^\eta}{w_0^\eta  e^{-\beta L_{t-1}^\eta}} \right)  = \ln\left(\frac{w_{t-1}^\eta}{w_t^\eta} \right) - \beta \ell_t(\x_t^\eta).
\end{equation}
Then, we have
\begin{equation} \label{eqn:w1}
\begin{split}
&\ln \left(\frac{W_t}{W_{t-1}} \right) = \ln \left(\frac{W_t}{W_{t-1}} \right) \sum_{\eta \in \H} w_t^\eta =  \sum_{\eta \in \H} w_t^\eta \ln \left(\frac{W_t}{W_{t-1}} \right) \\
\overset{(\ref{eqn:wt:wt1})}{=} &\sum_{\eta \in \H} w_t^\eta \ln\left(\frac{w_{t-1}^\eta}{w_t^\eta} \right) - \beta \sum_{\eta \in \H} w_t^\eta \ell_t(\x_t^\eta) \leq - \frac{1}{2} \|\w_t-\w_{t-1}\|_1^2 - \beta \sum_{\eta \in \H} w_t^\eta \ell_t(\x_t^\eta)
\end{split}
\end{equation}
where the last inequality is due to Pinsker's inequality \citep[Lemma 11.6.1]{Elements:IT}. Thus
\begin{equation} \label{eqn:WT}
\ln W_T= \ln W_0 +\sum_{t=1}^T\ln \left(\frac{W_t}{W_{t-1}}\right) \overset{(\ref{eqn:w1})}{=}\sum_{t=1}^T  \left(- \frac{1}{2} \|\w_t-\w_{t-1}\|_1^2 - \beta \sum_{\eta \in \H} w_t^\eta \ell_t(\x_t^\eta) \right).
\end{equation}

Combining (\ref{eqn:wt:w0}) with (\ref{eqn:WT}), we obtain
\[
- \beta \min_{\eta \in \H} \left(L_T^\eta + \frac{1}{\beta} \ln \frac{1}{w_0^\eta} \right) \leq \sum_{t=1}^T  \left(- \frac{1}{2} \|\w_t-\w_{t-1}\|_1^2 - \beta \sum_{\eta \in \H} w_t^\eta \ell_t(\x_t^\eta) \right)
\]
We complete the proof by rearranging the above inequality.
\section{Proof of Lemma~\ref{lem:lookhead:expert}}
The analysis is similar to that of Theorem 10 of \citet{SOCO:OBD}, which relies on a strong condition
\[
\x_t^\eta=\x_{t-1}^\eta - \eta \nabla f_t(\x_t^\eta).
\]
Note that the above equation is essentially the vanishing gradient condition of $\x_t^\eta$ when (\ref{eqn:lookhead:expert}) is unconstrained. In contrast, we only make use of the first-order optimality criterion of $\x_t^\eta$ \citep{Convex-Optimization}, i.e.,
\begin{equation} \label{eqn:greedy:1}
\left\langle \nabla f_t(\x_t^\eta) + \frac{1}{\eta} (\x_t^\eta-\x_{t-1}^\eta),  \y - \x_t^\eta \right \rangle \geq 0, \ \forall \y \in \X
\end{equation}
which is much weaker.

From the convexity of $f_t(\cdot)$, we have
\[
\begin{split}
 &f_t(\x_t^\eta)-f_t(\u_t) \\
\leq & \langle \nabla f_t(\x_t^\eta) ,  \x_t^\eta - \u_t \rangle \\
\overset{(\ref{eqn:greedy:1})}{\leq} & \frac{1}{\eta} \langle \x_t^\eta-\x_{t-1}^\eta,  \u_t  - \x_t^\eta \rangle = \frac{1}{2 \eta} \left( \|\x_{t-1}^\eta-\u_t \|^2 - \|\x_{t}^\eta-\u_t\|^2 - \|\x_t^\eta-\x_{t-1}^\eta\|^2\right) \\
= &\frac{1}{2 \eta} \left( \|\x_{t-1}^\eta-\u_{t-1} \|^2 - \|\x_{t}^\eta-\u_t\|^2 +  \|\x_{t-1}^\eta-\u_{t} \|^2 - \|\x_{t-1}^\eta-\u_{t-1} \|^2  - \|\x_t^\eta-\x_{t-1}^\eta\|^2\right) \\
=& \frac{1}{2 \eta} \left( \|\x_{t-1}^\eta-\u_{t-1} \|^2 - \|\x_{t}^\eta-\u_t\|^2 + \langle \x_{t-1}^\eta-\u_t + \x_{t-1}^\eta-\u_{t-1},\u_{t-1}- \u_t \rangle- \|\x_t^\eta-\x_{t-1}^\eta\|^2\right)  \\
\leq & \frac{1}{2 \eta} \left( \|\x_{t-1}^\eta-\u_{t-1} \|^2 - \|\x_{t}^\eta-\u_t\|^2 + \big(\| \x_{t-1}^\eta-\u_t \| +\| \x_{t-1}^\eta-\u_{t-1}\| \big) \|\u_t - \u_{t-1}\|\right)\\
& -  \frac{1}{2 \eta} \|\x_t^\eta-\x_{t-1}^\eta\|^2 \\
\overset{(\ref{eqn:domain})}{\leq}  &  \frac{1}{2 \eta} \left( \|\x_{t-1}^\eta-\u_{t-1} \|^2 - \|\x_{t}^\eta-\u_t\|^2 \right) + \frac{D}{\eta}\|\u_t - \u_{t-1}\| -  \frac{1}{2 \eta} \|\x_t^\eta-\x_{t-1}^\eta\|^2 .
\end{split}
\]

Summing the above inequality over all iterations, we have
\begin{equation} \label{eqn:greedy:2}
\begin{split}
\sum_{t=1}^T \left( f_t(\x_t^\eta) - f_t(\u_t) \right) \leq & \frac{1}{2 \eta}   \|\x_0^\eta-\u_0\|_2^2  + \frac{D}{\eta} \sum_{t=1}^T \|\u_t - \u_{t-1}\| -  \frac{1}{2 \eta} \sum_{t=1}^T \|\x_t^\eta-\x_{t-1}^\eta\|^2 \\
\overset{(\ref{eqn:domain})}{\leq} & \frac{1}{2 \eta}   D^2 + \frac{D}{\eta} \sum_{t=1}^T \|\u_t - \u_{t-1}\| -  \frac{1}{2 \eta} \sum_{t=1}^T \|\x_t^\eta-\x_{t-1}^\eta\|^2.
\end{split}
\end{equation}
Then, the dynamic regret with switching cost can be upper bounded as follows
\[
\begin{split}
&\sum_{t=1}^T \left( f_t(\x_t^\eta) +\|\x_t^\eta-\x_{t-1}^\eta\|- f_t(\u_t) \right) \\
\overset{(\ref{eqn:greedy:2})}{\leq} & \frac{1}{2 \eta}   D^2  + \frac{D}{\eta} \sum_{t=1}^T \|\u_t - \u_{t-1}\|  -  \frac{1}{2 \eta} \sum_{t=1}^T \|\x_t^\eta-\x_{t-1}^\eta\|^2 +\sum_{t=1}^T  \|\x_t^\eta-\x_{t-1}^\eta\|\\
\leq & \frac{1}{2 \eta}   D^2  + \frac{D}{\eta} \sum_{t=1}^T \|\u_t - \u_{t-1}\| -  \frac{1}{2 \eta} \sum_{t=1}^T \|\x_t^\eta-\x_{t-1}^\eta\|^2 +\sum_{t=1}^T  \left( \frac{1}{2 \eta} \|\x_t^\eta-\x_{t-1}^\eta\|^2 + \frac{\eta}{2} \right) \\
= &\frac{1}{2 \eta}   D^2  + \frac{D}{\eta} \sum_{t=1}^T \|\u_t - \u_{t-1}\|  + \frac{\eta T}{2}.
\end{split}
\]
\end{document}